%% file: main.tex
\pdfoutput=1
\documentclass{article}

\usepackage[main,final]{neurips_2026}

\usepackage[utf8]{inputenc}
\usepackage[T1]{fontenc}
\usepackage{amsmath,amssymb,amsthm,mathtools}
\usepackage{booktabs}
\usepackage{xcolor}
\usepackage{graphicx}
\usepackage{subcaption}
\usepackage{hyperref}
\usepackage{url}
\usepackage[expansion=false]{microtype}
\usepackage{float}

\usepackage{algorithm}
\usepackage{enumitem}

\newtheorem{theorem}{Theorem}[section]
\newtheorem{lemma}[theorem]{Lemma}
\newtheorem{proposition}[theorem]{Proposition}
\newtheorem{corollary}[theorem]{Corollary}

\newtheorem{assumption}{Assumption}
\newtheorem{remark}[theorem]{Remark}

\DeclareMathOperator*{\argmin}{arg\,min}

\DeclareMathOperator{\Var}{Var}
\DeclareMathOperator{\Cov}{Cov}

\newcommand{\R}{\mathbb{R}}
\newcommand{\E}{\mathbb{E}}

\newcommand{\cX}{\mathcal{X}}

\newcommand{\bm}[1]{\boldsymbol{#1}}
\newcommand{\blam}{\bm{\lambda}}
\newcommand{\bbeta}{\bm{\beta}}

\title{Your Hypergradient is Skewed: Antithetic Neumann Estimation for Bilevel Optimization}

\author{{\href{https://orcid.org/0000-0003-0800-966X}{Jason Bohne}}\thanks{{Jason Bohne and Pawe{\l} Polak contributed equally to this research.}}\\
  {Bloomberg}\\
  {New York City, NY, USA}\\
  {\makebox[\authorcolL]{\texttt{jbohne5@bloomberg.net}}}
  \And
  {\href{https://orcid.org/0000-0001-6645-015X}{Pawe{\l} Polak}}\textsuperscript{{*}}\thanks{{This research was conducted while participating in the Bloomberg Visiting Faculty Program during a sabbatical from Stony Brook University.}}\\
  {Stony Brook University, Stony Brook, NY, USA}\\
  {Bloomberg, New York City, NY, USA}\\
  {\makebox[\authorcolR]{\texttt{pawel.polak@stonybrook.edu}}}
  \AND
  {\href{https://orcid.org/0000-0003-0166-4338}{Gary Kazantsev}}\\
  {Bloomberg}\\
  {New York City, NY, USA}\\
  {\makebox[\authorcolL]{\texttt{gkazantsev@bloomberg.net}}}
  \And
  {\href{https://orcid.org/0000-0002-4593-9340}{David Rosenberg}}\\
  {Bloomberg}\\
  {Toronto, ON, Canada}\\
  {\makebox[\authorcolR]{\texttt{drosenberg44@bloomberg.net}}}
}

\hypersetup{pdfauthor={Jason Bohne, Paweł Polak, Gary Kazantsev, David Rosenberg},
  pdftitle={Your Hypergradient is Skewed: Antithetic Neumann Estimation for Bilevel Optimization},
  hidelinks}

\begin{document}

\newlength{\authorcolL}\settowidth{\authorcolL}{\normalfont\texttt{gkazantsev@bloomberg.net}}
\newlength{\authorcolR}\settowidth{\authorcolR}{\normalfont Stony Brook University, Stony Brook, NY, USA}
\maketitle

\begin{abstract}
 Bilevel optimization is a core tool in practical machine learning workflows, including hyperparameter tuning, data reweighting, robust learning, reinforcement learning, and imbalanced classification. Hypergradient computation requires the inverse of the inner Hessian, typically estimated via a truncated Neumann series. While existing analyses focus on bias and variance, we show that the product structure of the Neumann series introduces a third source of error: skewness. {We identify that, unlike ordinary gradients, whose asymmetry is suppressed by mini-batch averaging, the Neumann estimator} amplifies skewness across factors, with the third-moment error growing{, under weak contraction,} with truncation depth, inner condition number, and per-factor noise asymmetry. This arises in many practical settings such as corrupted labels, outliers, heterogeneous or {class-imbalanced} datasets. To address this, we introduce the Antithetic Neumann Estimator (ANE) which pairs each Hessian sample with a mirrored counterpart around a reference Hessian and averages the resulting inverse-Hessian estimates. With an exact reference the antithetic product retains only even-order noise, eliminating all odd-order {perturbation} terms; with an approximate reference, the residual odd-order contribution {is bounded linearly in} the reference error. ANE requires only a modification of the Hessian{--}vector step and combines with existing stochastic or variance-reduced bilevel methods. We instantiate ANE-SGD and ANE-VR, the latter adding recursive variance reduction. Across robust regression, {data cleaning}, offline reinforcement learning, and imbalanced classification, ANE-based methods {improve on stochastic and variance-reduced baselines on every task (20 seeds each), lowering loss by {4}--25\% relative to the best baseline; the largest gains, on robust regression, are a 25\% lower test loss and a 54\% higher test $R^2$}. {These results support} skewness reduction as a practical, complementary mechanism for improving hypergradient quality.
\end{abstract}

\section{Introduction}\label{sec:intro}

We study the bilevel optimization problem $\min_{\blam \in \cX} F(\blam) \triangleq f(\blam,\bbeta^*(\blam))$ with $\bbeta^*(\blam) \in \argmin_{\bbeta} g(\blam,\bbeta)$, where $\blam$ denotes the outer (upper-level) variables, $\bbeta^*(\blam)$ is the optimal inner (lower-level) solution, $f$ is the outer objective, and $g$ is the inner objective. This formulation underlies hyperparameter optimization \citep{franceschi2017forward,maclaurin2015gradient,lorraine2020optimizing}, meta-learning \citep{finn2017model,rajeswaran2019meta}, and data reweighting \citep{ren2018learning,shu2019meta}. Gradient-based methods optimize $F$ using the \emph{hypergradient} $\nabla F(\blam)$, which requires differentiating through the inner solution $\bbeta^*(\blam)$. The standard scalable approach approximates the resulting inverse Hessian via a truncated Neumann series \citep{ghadimi2018approximation}. A substantial literature has since developed variance-reduced and oracle-efficient bilevel methods \citep{ji2021bilevel,yang2021provably,khanduri2021near,dagreou2022framework,hong2023two,chen2024optimal,dagreou2024lower}, but these works evaluate hypergradient estimators {primarily} through their bias and variance.

The Neumann estimator contains products of stochastic Hessian factors. In a scalar product model, asymmetric per-factor noise appears in odd perturbation-degree terms. These terms can make unnormalized third-moment proxies large when contraction is weak, as occurs along poorly conditioned directions, even though the standardized scalar skewness scales like $1/\sqrt m$ in the perturbative regime (Theorem~\ref{thm:skew_scalar}). This third-order mechanism is present in {Neumann-based bilevel methods such as} StocBiO~\citep{ji2021bilevel} {and} SUSTAIN~\citep{khanduri2021near} whenever the stochastic Hessian noise is asymmetric. Existing variance-reduced bilevel methods~\citep{khanduri2021near,dagreou2022framework} reduce across-iteration variance but act at a different stage than the within-estimate Neumann product.

To address this, we introduce the \emph{antithetic Neumann estimator} (ANE). ANE pairs each stochastic Hessian sample with a mirrored factor around a reference Hessian $\bar H$ and averages the two products: when $\bar H$ equals the true Hessian, all odd-order noise terms, {including} those carrying $\mu_3'$, cancel. We further embed ANE inside a recursive variance-reduction outer loop, yielding ANE-VR.

\begin{table}[t]
\centering
\small
\caption{\textbf{Neumann-based bilevel methods.} All methods inherit truncation bias from the Neumann inverse approximation. SUSTAIN applies recursive variance reduction across outer iterations. ANE modifies the within-estimate Neumann product by antithetic mirroring, which cancels odd perturbation-degree terms under an exact reference. These mechanisms act at different stages and can be combined; ANE may also affect variance because it removes the linear perturbation term.}
\label{tab:method_comparison}
\begin{tabular}{@{}lccc@{}}
\toprule
Algorithm & \shortstack{Truncation\\bias} & \shortstack{Recursive /\\across-iter. VR} & \shortstack{Odd-degree\\cancellation} \\
\midrule
StocBiO~\citep{ji2021bilevel}    & $\checkmark$ & $\times$     & $\times$     \\
SUSTAIN~\citep{khanduri2021near} & $\checkmark$ & $\checkmark$ & $\times$     \\
ANE-SGD \emph{(ours)}            & $\checkmark$ & $\times$     & $\checkmark$ \\
ANE-VR \emph{(ours)}             & $\checkmark$ & $\checkmark$ & $\checkmark$ \\
\bottomrule
\end{tabular}
\end{table}

\begin{figure*}[t]
\centering
\includegraphics[width=0.85\textwidth]{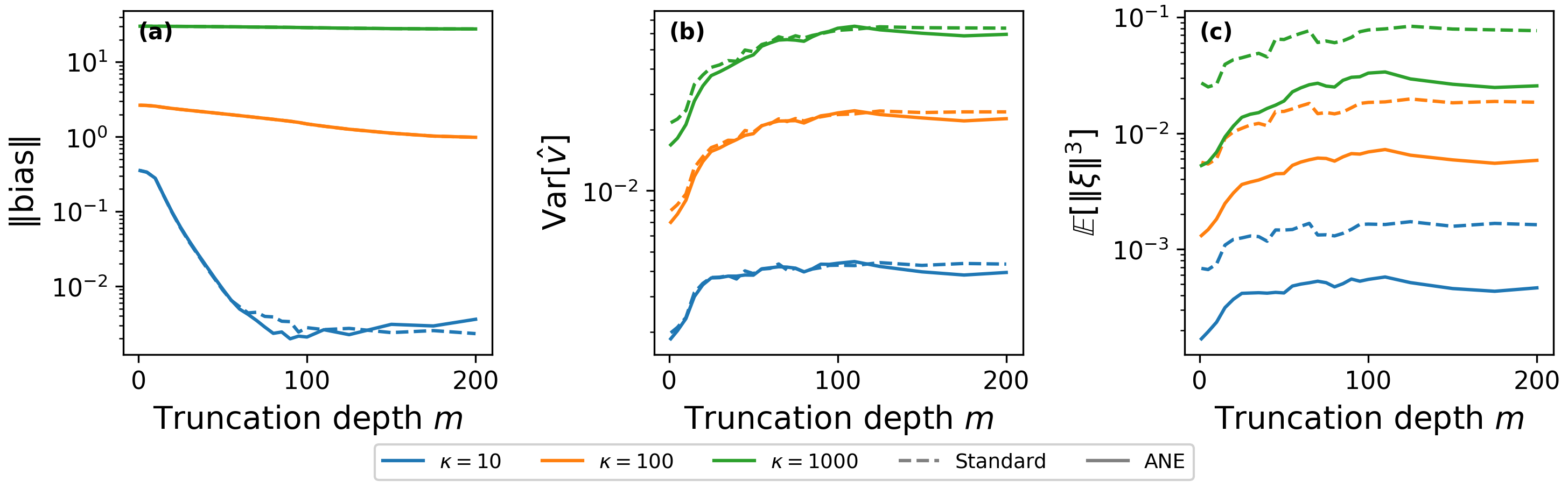}
\caption{\textbf{Synthetic illustration of the ANE mechanism} (see {the} synthetic experiment in Section~\ref{sec:experiments}). Controlled quadratic setting $g(\blam,\bbeta)=\tfrac12\bbeta^\top H\bbeta-\blam^\top C\bbeta$, prescribed $\kappa_g\in\{10,100,1000\}$, with stochastic {Hessian--vector products (HVPs)} $\hat H v=s_i\cdot Hv$, $\E[s_i]=1$, positive third central moment ($d=50$; dashed: Standard, solid: ANE). ANE and Standard share the same deterministic truncation target. (a)~Bias: matched. (b)~Variance: empirically comparable. (c)~Absolute third-moment proxy $\E[\|\xi\|^3]$ {(centered, i.e.\ $\mathcal M_3(\xi)=\E\|\xi-\E\xi\|^3$; a scale-sensitive vector diagnostic, not the signed skewness $\gamma_1$ of Theorem~\ref{thm:skew_scalar})}: lower for ANE ($41$--$74\%$ reduction across the swept $\kappa_g${, relative to a single standard estimate}), consistent with odd perturbation-degree cancellation. The variance comparison is empirical for this setup, not a general consequence of the theory.}
\label{fig:teaser}\label{fig:synthetic_diagnostics}
\end{figure*}

Throughout, $\gamma_1(Y)=\E[(Y-\E Y)^3]/\Var(Y)^{3/2}$ is signed standardized skewness for scalar $Y$, and $\mathcal M_3(\xi)=\E\|\xi-\E\xi\|^3$ is a scale-sensitive absolute third-moment proxy for vector error $\xi$ (upper-bounding magnitudes of directional signed third moments, but itself \emph{not} a signed skewness). {The centered absolute moment can remain nonzero even for a symmetric error distribution.}

\paragraph{Contributions.}
\begin{enumerate}[leftmargin=*,labelindent=0pt,itemsep=2pt,topsep=2pt]
  \item We isolate a third-order mechanism in stochastic Neumann products: in a scalar model, the signed standardized skewness depends on the per-factor third central moment, while unnormalized third-moment contributions can be large under weak contraction (Theorem~\ref{thm:skew_scalar}); a coordinate bound lifts scalar absolute third-moment effects to the vector inverse-Hessian estimate (Proposition~\ref{prop:vector_third}).
{Under weak contraction, we further give lower bounds for the full truncated sum showing that, in this model, the vector proxy grows with truncation depth and conditioning (Lemma~\ref{lem:sum_growth}).}
  \item We introduce ANE, an antithetic Neumann estimator that mirrors stochastic Hessian factors around a reference $\bar H$. With {an} exact reference and $\tau=1$, ANE cancels all odd perturbation-degree terms in each product (Proposition~\ref{prop:odd_cancel}); under uniform contraction, the deviation from the exact-reference estimator is {bounded linearly in} the reference error (Proposition~\ref{prop:reference_error}).
{The cancellation holds for noncommuting matrix-valued perturbations and does not require a shared eigenbasis.}
{An iterate-level convergence bound (Theorem~\ref{thm:iterate}) shows the benefit: {for Hessian error at level $\rho<1$} at fixed depth, ANE shrinks the bound's inverse-Hessian third-moment term from {$O(\rho^3)$ to $O(\rho^6)$} (Proposition~\ref{prop:moment_scaling}).}
  \item We instantiate ANE-SGD and ANE-VR as modifications of the inverse-Hessian--vector primitive. Recursive variance reduction and ANE act at different stages of the estimator, although ANE may also affect variance. Experiments on four bilevel tasks (robust regression, data cleaning, offline RL, imbalanced classification) show ANE-based variants improve the reported metrics, consistent with the proposed odd-order cancellation mechanism{; matched synthetic controls isolate it, with no gain under symmetric noise, and runtime is measured at practical depths}.
\end{enumerate}

\section{Background and Related Work}\label{sec:prelim}

\paragraph{Assumptions.}
We adopt the standard assumptions for Neumann-based bilevel optimization~\citep{ghadimi2018approximation,ji2021bilevel,khanduri2021near} (Assumptions~\ref{ass:smooth}--\ref{ass:stoch}), plus one additional assumption that goes beyond the existing literature (Assumption~\ref{ass:higher}).

\begin{assumption}[Smoothness]\label{ass:smooth}
$f$ and $g$ are continuously differentiable; $g$ is twice differentiable in $\bbeta$. The maps $\nabla f$, $\nabla_{\bbeta} g$, $\nabla^2_{\blam\bbeta}g$, and $\nabla^2_{\bbeta\bbeta}g$ are Lipschitz continuous with respective constants $\ell_{f,1}$, $\ell_{g,1}$, $\ell_{g,\blam\bbeta}$, and $\ell_{g,2}$.
\end{assumption}
\begin{assumption}[Strong convexity]\label{ass:sc}
For every $\blam \in \cX$, $g(\blam,\cdot)$ is $\mu_g$-strongly convex. The inner condition number is $\kappa_g \triangleq \ell_{g,1}/\mu_g$.
\end{assumption}
\begin{assumption}[Bounded stochastic oracles]\label{ass:stoch}
Stochastic gradient, Jacobian, and Hessian estimators are unbiased with uniformly bounded second moments: $\E[\|\widehat{H} - H\|^2] \leq \sigma_H^2$ (and analogously for gradients and Jacobians with constants $\sigma_g^2$, $\sigma_J^2$).
\end{assumption}
\begin{assumption}[Higher-order regularity]\label{ass:higher}
(i)~\emph{Bounded third moment.} The stochastic Hessian estimator satisfies $\E[\|\widehat{H} - H\|^3] \leq {(\hat\eta\rho)^3}${, i.e., the scaled error $E\triangleq(\widehat H-H)/\hat\eta$ has $(\E\|E\|^3)^{1/3}\le\rho$. The regime $\rho<1$ is that of the standard estimator itself: with per-sample Hessians in $[0,\ell_{g,1}I]$ and the step size $\hat\eta>\ell_{g,1}$ of Section~\ref{sec:prelim}, $\|E\|\le\ell_{g,1}/\hat\eta<1$}.
  (ii)~\emph{{Outer regularity}.}   {For the iterate-level bound, assume directly that $F$ is bounded below by $F_\star$, has $L_F$-Lipschitz gradient, and has $L_2$-Lipschitz Hessian on a domain containing the update segments. These properties are additional requirements: they do not follow from Assumptions~\ref{ass:smooth}--\ref{ass:stoch} alone, and a Lipschitz $\nabla^2F$ requires, for example, Lipschitz third derivatives of $g$.}
 \end{assumption}

Assumptions~\ref{ass:smooth}--\ref{ass:stoch} are standard~\citep{ghadimi2018approximation,ji2021bilevel,khanduri2021near}. {Assumption~\ref{ass:higher} collects the additional third-moment condition used {with Theorem~\ref{thm:iterate} (Proposition~\ref{prop:moment_scaling}; Corollary~\ref{cor:quadratic_bound} uses its almost-sure form)} and the outer regularity conditions used in Proposition~\ref{prop:outer_descent}.} {The analysis takes $\hat\eta>\ell_{g,1}$; the experiments use the practical choice $\hat\eta=\hat L_{\max}/2$ (Table~\ref{tab:hyperparams}), for which $\rho<1$ is not guaranteed, with safeguards keeping the recursions bounded (Appendix~\ref{app:real_diagnostics}).} Some algebraic results below impose additional local assumptions stated with each result: the scalar perturbative expansion (Theorem~\ref{thm:skew_scalar}) requires a finite sixth moment; Corollary~\ref{cor:quadratic_bound} assumes an almost-sure operator-norm bound on the Hessian perturbations; Proposition~\ref{prop:reference_error} assumes uniform contraction of both mirrored branches.

Under Assumptions~\ref{ass:smooth}--\ref{ass:sc}, $\bbeta^*(\blam)$ is unique and the implicit hypergradient is $\nabla F(\blam)
  = \nabla_{\blam} f - \nabla^2_{\blam\bbeta} g\;
    \bigl(\nabla^2_{\bbeta\bbeta} g\bigr)^{-1}
    \nabla_{\bbeta} f,$ evaluated at $(\blam,\bbeta^*(\blam))$~\citep{ghadimi2018approximation}. The bottleneck is the inverse-Hessian--vector product $(\nabla^2_{\bbeta\bbeta}g)^{-1}\nabla_{\bbeta} f$, approximated via a truncated Neumann series. Setting $\hat A_j \triangleq I - \widehat{H}_j/\hat\eta$ for i.i.d.\ Hessian samples and $\hat\eta > \ell_{g,1}$:
\begin{equation}\label{eq:stoch_neumann}
  \widehat{v}_{\rm Neu}
  = \frac{1}{\hat{\eta}}\sum_{k=0}^{m-1}\prod_{j=1}^{k} \widehat{A}_j\, b,
  \qquad
  \widetilde{\nabla} F(\blam)
  = \widehat{\nabla}_{\blam} f - \widehat{\nabla}^2_{\blam\bbeta} g\;\widehat{v}_{\rm Neu}.
\end{equation}
Truncation depth $m = O(\kappa_g{\log(1/\epsilon)})$ controls bias {to accuracy $\epsilon$}~\citep{ghadimi2018approximation}. This stochastic hypergradient is shared by StocBiO~\citep{ji2021bilevel} and SUSTAIN~\citep{khanduri2021near}.

\paragraph{Bilevel optimization.}
Neumann-based implicit differentiation for bilevel optimization was introduced by \citet{ghadimi2018approximation}, with practical {applications} in hyperparameter optimization~\citep{pedregosa2016hyperparameter,maclaurin2015gradient,lorraine2020optimizing}, meta-learning~\citep{finn2017model,rajeswaran2019meta}, and data reweighting~\citep{ren2018learning,shu2019meta}. Subsequent work developed single-loop, two-timescale, and variance-reduced algorithms with improved oracle complexity~\citep{ji2021bilevel,yang2021provably,khanduri2021near,dagreou2022framework,hong2023two,arbel2022amortized}, including near-optimal lower bounds~\citep{chen2024optimal,dagreou2024lower}. These analyses characterize bias and variance of the hypergradient estimator.

\paragraph{Variance reduction.}
Recursive variance reduction via SARAH~\citep{nguyen2017sarah}, SPIDER~\citep{fang2018spider}, PAGE~\citep{li2021page}, and momentum-based methods~\citep{cutkosky2019momentum} has been widely adopted in bilevel optimization~\citep{khanduri2021near,dagreou2022framework}. These methods operate on the hypergradient \emph{after} the Neumann product has been formed, reducing the across-iteration variance.

\paragraph{Antithetic variates and adjacent work.}
Classical antithetic variates~\citep{hammersley1964monte} pair scalar or vector samples to cancel the linear noise term. Alternative inverse approximations~\citep{prakash2025bilevel} and heavy-tailed bilevel methods~\citep{nguyen2025stochastic,chen2025nonconvex} address different aspects of robustness. Online  bilevel formulations~\citep{tarzanagh2024online,bohne2024bregman,bohne2025smoothfbo} also share the Neumann primitive.

 \section{\texorpdfstring{{Skewness in Neumann Hypergradients and Its Reduction}}{Skewness in Neumann Hypergradients and Its Reduction}}\label{sec:skewness}

A mini-batch gradient is {an average} of i.i.d. contributions, so $\gamma_1$ decreases with batch size. The Neumann inverse approximation is structurally different: it is a \emph{sum of products} of random Hessian factors. Those products do not admit term-wise averaging; their third-order behavior is governed by multiplicative interactions between perturbations.
We work with the truncated-sum Neumann estimator {$\hat v_{\rm Neu}$ of~\eqref{eq:stoch_neumann}}. We build the analysis bottom-up: a scalar product model isolating the multiplicative mechanism (Section~\ref{sec:scalar_model}), its coordinate-wise lift (Section~\ref{sec:multivariate}), a decomposition of the full hypergradient error (Section~\ref{sec:full_chain}), and finally the antithetic construction that cancels odd perturbation-degree terms (Section~\ref{sec:ane}).

\subsection{Scalar Product Model}\label{sec:scalar_model}

As a stylized model, suppose the stochastic Hessian acts multiplicatively along an eigendirection $v$ of $H$: for i.i.d. scalar multipliers $s_j$ with $\E[s_j]=1$, $\hat H_jv=s_jHv$. Then each Neumann factor shares the eigenbasis of $H$, and along eigendirection $i$ with eigenvalue $\mu_i$ the factor reduces to a scalar variable $X_j^{(i)}=1-s_j\mu_i/\hat\eta=a_i+\delta_j^{(i)}$, with $a_i\triangleq1-\mu_i/\hat\eta$. This motivates the following scalar product model.

\begin{theorem}[Scalar product third moments]\label{thm:skew_scalar}
Let $X_1,\ldots,X_m$ be i.i.d. copies of $X=a+\delta$ with $0<|a|<1$, $\E\delta=0$, $\E\delta^2=\sigma^2$, $\E\delta^3=\mu_3'$, and $\E|\delta|^6<\infty$, and define $P_m\triangleq\prod_{j=1}^m X_j$. Then:
\begin{enumerate}
  \item[(i)] $\E[P_m]=a^m$ and $\Var(P_m)=(a^2+\sigma^2)^m-a^{2m}$.
  \item[(ii)] The third central moment is
  \begin{equation}\label{eq:mu3_product}
    \E[(P_m-a^m)^3]=(a^3+3a\sigma^2+\mu_3')^m-3(a^2+\sigma^2)^m a^m+2a^{3m}.
  \end{equation}
  \item[(iii)] In the perturbative regime $m\sigma^2/a^2=o(1)$ with $|a|$ bounded away from zero and $|\mu_3'|\le C\sigma^3$, $\Var(P_m)=m\sigma^2a^{2m-2}+O(m^2\sigma^4|a|^{2m-4})$ and $\E[(P_m-a^m)^3]=m\mu_3'a^{3m-3}+O(m^2\sigma^4|a|^{3m-4})$, and therefore the signed standardized skewness satisfies $\gamma_1(P_m)=\operatorname{sgn}(a)^{m-1}\mu_3'/(\sigma^3\sqrt m)+O(\sqrt m\,\sigma)$. For $a>0$ the sign factor is unity, recovering the positive-factor formula.
\end{enumerate}
\end{theorem}

The scalar model shows that odd perturbation-degree terms are proportional to the per-factor third central moment. Their unnormalized contribution $m\mu_3'a^{3m-3}$ is most pronounced under weak contraction, when $|a|$ is close to one, whereas the signed standardized skewness scales as $1/\sqrt m$ in the perturbative regime; for fixed $|a|<1$, the unnormalized contribution decays exponentially in $m$.

\subsection{From Scalar to Vector}\label{sec:multivariate}

The Neumann estimator error is vector-valued. In the eigenbasis of $H$, each product term decomposes into scalar factors of the form analyzed above.
{The scalar formulas require the multiplicative eigendirection model; the odd-degree cancellation and reference-sensitivity results below apply to general matrix-valued noise and do not require a shared eigenbasis.}

\begin{proposition}[Coordinate contribution to vector absolute third moments]\label{prop:vector_third}
Under the multiplicative eigendirection model, let the depth-$k$ product error along eigendirection $i$ be $\xi_i^{(k)}\triangleq(b_i/\hat\eta)(P_k^{(i)}-a_i^k)$, with $P_k^{(i)}\triangleq\prod_{j=1}^k(a_i+\delta_j^{(i)})$ and $b_i$ the $i$th coordinate of $b$ in the eigenbasis of $H$. Then
\[
  \E[\|\xi^{(k)}\|^3]\ge \max_i\frac{|b_i|^3}{\hat\eta^3}\E[|P_k^{(i)}-a_i^k|^3],
\]
so a large absolute third centered moment in any eigendirection lower-bounds the vector absolute third-moment proxy.
\end{proposition}

The proposition is a lower bound on $\mathcal M_3(\xi^{(k)})$: it shows that one-dimensional absolute third-moment effects survive in the multivariate proxy. It does not assert monotonic growth in $k$, nor does it bound signed vector skewness.
{Appendix~\ref{app:sum_growth} lower-bounds the full truncated sum: under weak contraction, $\mathcal M_3$ grows at least as $\sigma^3m^{9/2}$ for either sign of the factor-noise third moment, and as $\kappa_g^{3/2}$ along the smallest eigenvalue when $m=\Theta(\kappa_g)$; when that third moment {$\mu_3'$} is nonnegative, the signed third moment grows at least as {$\mu_3'm^4$}.}

\subsection{Full Hypergradient Error: Where the $m$-Dependence Enters}\label{sec:full_chain}

Let $G\triangleq\nabla_{\blam}f-Jv^\star$ with $v^\star\triangleq H^{-1}\nabla_{\bbeta}f$ and $J\triangleq\nabla^2_{\blam\bbeta}g$, and define the stochastic hypergradient $\widehat G\triangleq\widehat\nabla_{\blam}f-\widehat J\hat v$, where $\hat v$ is any stochastic inverse-Hessian{--}vector estimator (standard Neumann or ANE).

\begin{proposition}[Full-hypergradient error decomposition]\label{prop:skew_budget}
The stochastic hypergradient error admits the exact decomposition
\begin{equation}\label{eq:full_hg_error_decomp}
  \widehat G-G=(\widehat\nabla_{\blam}f-\nabla_{\blam}f)-(\widehat J-J)\hat v-J(\hat v-v^\star).
\end{equation}
Consequently,
\begin{equation}\label{eq:full_hg_third_bound}
  \|\widehat G-G\|^3\le 9\|\widehat\nabla_{\blam}f-\nabla_{\blam}f\|^3+9\|(\widehat J-J)\hat v\|^3+9\|J\|^3\|\hat v-v^\star\|^3.
\end{equation}
In particular, the explicit Neumann product and truncation-depth dependence enter through the inverse-Hessian--vector estimator $\hat v$; the outer-gradient and Jacobian oracles do not themselves contain Neumann products.
\end{proposition}

This motivates targeting the Neumann product: although other stochastic oracle terms may contribute to the full hypergradient error, the product expansion and its truncation-depth dependence arise inside $\hat v$.
We use the Neumann-only absolute third-moment proxy
\begin{equation}\label{eq:m3_diag}
  \mathcal M_3(\xi)\triangleq\E[\|\xi-\E\xi\|^3],\qquad \xi\triangleq\hat v_{\rm Neu}-v^\star,
\end{equation}
as a scale-sensitive diagnostic. It upper-bounds the magnitude of every directional signed third moment since $|\E[(u^\top(\xi-\E\xi))^3]|\le\E[\|\xi-\E\xi\|^3]$ for every unit $u$, but $\mathcal M_3$ is itself \emph{not} a signed skewness and can be nonzero for symmetric distributions.

\subsection{The Antithetic Neumann Estimator}\label{sec:ane}

For each stochastic Hessian $\hat H_j$, ANE constructs a mirror around a reference Hessian $\bar H$. Let $\tau\in[0,1]$ be a mirror strength and define
\begin{equation}\label{eq:mirror_hessian}
  \hat H_j^+\triangleq\hat H_j,\qquad \hat H_j^-\triangleq\bar H-\tau(\hat H_j-\bar H).
\end{equation}
Setting $\hat A_j^\pm\triangleq I-\hat H_j^\pm/\hat\eta$, the antithetic Neumann estimator (ANE) is
\begin{equation}\label{eq:ane_operator}
  \hat v_{\rm ANE}\triangleq\frac{1}{2\hat\eta}\sum_{k=0}^{m-1}\left[\prod_{j=1}^k\hat A_j^++\prod_{j=1}^k\hat A_j^-\right]b.
\end{equation}
With {an} exact reference and $\tau=1$, the average of forward and mirrored products cancels every term containing an odd number of perturbation factors (algebraic cancellation in the product expansion; this does not imply zero signed or absolute third moments overall). For $\tau<1$ the linear term is not fully canceled, with residual $\propto 1-\tau$ in the scalar expansion. The full ANE hypergradient replaces $\hat v_{\rm Neu}$ by $\hat v_{\rm ANE}$; {in practice} $\bar H$ can be maintained by periodic snapshots {on a larger batch (four times the HVP batch in our experiments)}.
{The reference can be represented by its Hessian--vector-product operator, so a dense $p\times p$ matrix need not be stored.}

\paragraph{Why does this cancel odd-degree terms?}
With two scalar factors, $\tfrac12[(a+\delta_1)(a+\delta_2)+(a-\delta_1)(a-\delta_2)]=a^2+\delta_1\delta_2$, so the linear terms cancel. The same mechanism holds at arbitrary depth.

\begin{proposition}[Odd perturbation-degree cancellation]\label{prop:odd_cancel}
Assume $\bar H=H$ and $\tau=1$. Define $A\triangleq I-H/\hat\eta$ and $E_j\triangleq(\hat H_j-H)/\hat\eta$, so that $\hat A_j^+=A-E_j$ and $\hat A_j^-=A+E_j$. Then at each depth $k$,
\begin{equation}\label{eq:even_order_expansion}
  \frac12\left(\prod_{j=1}^k(A-E_j)+\prod_{j=1}^k(A+E_j)\right)=\sum_{\substack{S\subseteq[k]\\|S|\ \mathrm{even}}}\Pi_k(S),
\end{equation}
where $\Pi_k(S)$ is the ordered product with factor $E_j$ for $j\in S$ and factor $A$ otherwise. All terms with an odd number of perturbation factors vanish.
\end{proposition}

\begin{corollary}[Quadratic-in-$\rho$ perturbation bound]\label{cor:quadratic_bound}
Assume the setting of Proposition~\ref{prop:odd_cancel} and that $\|A\|\le a<1$ and $\|E_j\|\le\rho$ almost surely for all $j${ (the almost-sure form of Assumption~\ref{ass:higher}(i))}.
 Then for small $\rho$,
\begin{equation}\label{eq:quadratic_bound}
\begin{split}
  \|\hat v_{\rm ANE}-\mathcal N_m(H;b)\|
  &\le \frac{\|b\|\rho^2}{2\hat\eta}\sum_{k=0}^{m-1}k(k-1)a^{k-2}\\
  &\quad+O\!\left(\frac{\|b\|\rho^4}{\hat\eta}\sum_{k=0}^{m-1}k^4a^{k-4}\right),
\end{split}
\end{equation}
where $\mathcal N_m(H;b)=\hat\eta^{-1}\sum_{k=0}^{m-1}A^kb$. Consequently, if $a\le a_0<1$ is bounded away from one, then $\|\hat v_{\rm ANE}-\mathcal N_m(H;b)\|=O(\rho^2\|b\|/[\hat\eta(1-a_0)^3])$ uniformly in $m$. In the weak-contraction finite-depth regime $a\approx1$, the quadratic term scales at most as $O(m^3\rho^2\|b\|/\hat\eta)$. In both cases the leading perturbation order in $\rho$ is quadratic, whereas the standard truncated Neumann operator has a generically linear perturbation term.
\end{corollary}

Thus ANE removes the leading odd perturbation-degree terms when the reference is exact. This is the algebraic mechanism behind {the smaller inverse-Hessian third-moment bound of Proposition~\ref{prop:moment_scaling}}.

\begin{proposition}[Sensitivity to approximate reference]\label{prop:reference_error}
The following is a local stability bound; it assumes both the exact-reference and approximate-reference mirrored factors are uniformly contractive (this is stronger than the moment assumptions on stochastic Hessian noise). Let $R=\bar H-H$ and $B=2R/\hat\eta$. Assume $\tau=1$ and $\|A+E_j\|,\|A-B+E_j\|\le c<1$ for all $j$. Then
\begin{equation}\label{eq:reference_error_bound}
  \|\hat v_{\rm ANE}(\bar H)-\hat v_{\rm ANE}(H)\|\le\frac{\|B\|\|b\|}{2\hat\eta}\sum_{k=1}^{m-1}kc^{k-1}\le\frac{\|B\|\|b\|}{2\hat\eta(1-c)^2}.
\end{equation}
An approximate reference shifts the mean of the mirrored branch ($\E[\hat H_j^-]=H+2R$ when $\bar H=H+R$ and $\tau=1$), so Proposition~\ref{prop:reference_error} is a sensitivity bound: under uniform contraction, the approximate-reference estimator differs from the exact-reference estimator by $O(\|R\|)$, controlling residual odd-degree terms and bias from a stale reference.
\end{proposition}

\begin{remark}[Practical stabilization]\label{rem:stabilization}
When $2\bar H-\hat H_j$ becomes poorly conditioned, ANE can be stabilized by damped mirroring ($\tau<1$), periodic refreshes of $\bar H$, or clipping in operator norm or Hessian--vector-product norm. These are stabilization choices that trade exact cancellation for robustness: the algebraic odd-degree cancellation in Proposition~\ref{prop:odd_cancel} applies exactly only when $\tau=1$ and the reference is exact.
{With an exact reference, a degree-$n$ perturbation term has coefficient $[1+(-\tau)^n]/2$ relative to Standard. Thus odd terms retain $(1-\tau^n)/2$: $\tau=0.5$ leaves one quarter of the leading linear coefficient, while $\tau=1$ removes it. At $\tau=0$, ANE averages the Standard and reference branches rather than reducing to Standard.}
\end{remark}

\paragraph{Relation to variance reduction.}
Although ANE is motivated by odd-degree cancellation, the same algebra also removes the linear perturbation term in the product expansion under an exact reference; ANE can therefore affect variance as well as third moments. Its complementarity with recursive variance reduction is algorithmic: ANE modifies the within-estimate Neumann product, whereas recursive variance reduction reduces across-iteration fluctuations of the hypergradient estimator.

\section{Algorithms}\label{sec:ane_vr}

\paragraph{ANE-SGD.}
The simplest use of ANE is as a drop-in replacement for the standard Neumann inverse approximation inside a bilevel method such as StocBiO~\citep{ji2021bilevel}. We call the resulting method \textbf{ANE-SGD}. The inner solver, outer step size, and the surrounding single-loop structure are left unchanged; the only modification is the inverse-Hessian{--}vector estimate, where each stochastic Hessian factor is paired with its mirror around the current reference $\bar H_t$. Comparing a standard Neumann implementation against ANE-SGD isolates the effect of the antithetic estimator without the additional confound of recursive outer variance reduction.

 \paragraph{ANE-VR.}
ANE targets the within-estimate asymmetry of the Neumann inverse approximation, while recursive variance reduction targets the across-iteration fluctuation of the full hypergradient. Since these affect different parts of the estimation pipeline, they compose naturally. Replacing the Neumann inverse approximation inside a {PAGE-type recursive} wrapper{~\citep{li2021page}} yields \textbf{ANE-VR} (Algorithm~\ref{alg:ane_vr}).
{Full algorithms appear in Appendix~\ref{app:algorithms}; Algorithm~\ref{alg:ane_vr_compact} gives the compact PAGE variant. ANE-SGD is the simpler default, while ANE-VR is useful in an existing recursive-VR pipeline or when substantial across-iteration variance remains.}

\paragraph{{Computational cost.}}
{ANE evaluates forward and mirrored branches and also applies the reference operator. In a matrix-free implementation, a mirrored HVP uses $\hat H^-v=(1+\tau)\bar Hv-\tau\hat Hv$ at the mirrored branch vector, which generally differs from the forward vector, so each Neumann factor costs three HVP-type operations (one stochastic HVP per branch and one reference HVP) against one for Standard. Refreshing the reference every $q_{\rm ref}=10$ steps only resamples its batch (four times the HVP batch) and is negligible; applying it at every mirrored factor is included in the measured hypergradient cost (Appendix~\ref{app:timing_sweep}). Storage can remain linear in parameter count; computational scaling also depends on architecture and reference-batch size.
}

\begingroup
\begin{theorem}[Iterate-level guarantee]\label{thm:iterate}
{Let $F$ be bounded below by $F_\star$, with $L_F$-Lipschitz gradient and $L_2$-Lipschitz Hessian, and let $\blam_{t+1}=\blam_t-\alpha d_t$ with $0<\alpha\le1/(4L_F)$. Let $\mathcal F_t$ contain $\blam_t$ and all randomness preceding $d_t$, write $e_t=d_t-\E[d_t\mid\mathcal F_t]$ and $y_t=\blam_t-\alpha\E[d_t\mid\mathcal F_t]$, and assume that, almost surely for all $t$,
}\[
\begin{gathered}
\|\E[d_t\mid\mathcal F_t]-\nabla F(\blam_t)\|^2\le\bar r^2,\qquad
\E[\|e_t\|^2\mid\mathcal F_t]\le\bar\sigma^2,\\
\E[\|e_t\|^3\mid\mathcal F_t]\le\bar M_3,\qquad
\E[e_t^\top\nabla^2F(y_t)e_t\mid\mathcal F_t]\le\bar Q.
\end{gathered}
\]
{Then
}\begin{equation}\label{eq:outer_rate_main}
 \frac{1}T\sum_{t<T}\E\|\nabla F(\blam_t)\|^2
 \le \frac{2(F(\blam_0)-F_\star)}{\alpha T}
 +\frac{5}{2}\bar r^2+\alpha\bar Q
 +\frac{L_2\alpha^2}{3}\bar M_3.
\end{equation}
\end{theorem}
{The proof is in Appendix~\ref{app:outer_descent}. Since $\nabla^2F\preceq L_FI$, taking $\bar Q=L_F\bar\sigma^2$ is always valid and gives the standard first-order variance term $L_F\alpha\bar\sigma^2$; the cubic term is the price of using the exact curvature, which pays off when the fluctuations $e_t$ lie along low-curvature directions ($\bar Q<L_F\bar\sigma^2$). {For $d_t=\widehat G$ with the Hessian error at level $\rho$ (Assumption~\ref{ass:higher}(i)), the decomposition of Proposition~\ref{prop:skew_budget}, applied to the centered error, splits $\bar M_3$ into an oracle part $\bar M_{\rm orc}$ from the outer-gradient and Jacobian noise, bounded alike for both estimators (Appendix~\ref{app:moment_scaling}), and the inverse-Hessian part $9\|J\|^3\E\|\hat v-\mathcal N_m(H;b)\|^3$. At fixed depth, with an exact reference and $\tau=1$ (Proposition~\ref{prop:moment_scaling}),
\[
\bar M_3\le\bar M_{\rm orc}+\begin{cases}O(\rho^3)&\text{standard Neumann},\\ O(\rho^6)&\text{ANE}.\end{cases}
\]
{Before its constants are collected, ANE's bound keeps only the even-order terms of the standard bound, so it is never larger (Appendix~\ref{app:moment_scaling}).}}
}
\endgroup

\section{Experiments}\label{sec:experiments}

\begin{algorithm}[t]
 \caption{ANE-VR (compact): ANE-SGD with recursive outer variance reduction}
\label{alg:ane_vr_compact}
\begin{flushleft}
\textbf{Input:} Same as Algorithm~\ref{alg:ane_sgd}, plus snapshot probability $p$.\\[3pt]
\textbf{Initialize:} $\blam_0$, $\bbeta_0$, $\bar H_0${, and $d_{-1}=0$.} \\[3pt]
\textbf{for} $t=0,1,\ldots,T-1$ \textbf{do}
\begin{enumerate}[leftmargin=1.5em]
  \item Steps 1--3 of Algorithm~\ref{alg:ane_sgd} $\to$ $\widehat G_t^{\mathrm{ANE}}$.

  \item \textbf{Variance reduction (PAGE).}\\
  (a) {At $t=0$ or with} probability $p$: $d_t \leftarrow \widehat G_t^{\mathrm{ANE}}$ {(snapshot from the current mini-batch)}.\\
  (b) Otherwise: {$d_t \leftarrow d_{t-1} + \widehat G_t^{\mathrm{ANE}} - \widehat G^{\mathrm{ANE}}(\blam_{t-1},\bbeta_{t-1})$, where the second term is recomputed at the previous iterate on fresh samples with the current reference.}

  \item \textbf{Outer step.}
  $\blam_{t+1} \leftarrow \blam_t - \alpha\, d_t$.
\end{enumerate}
\textbf{end for}
\end{flushleft}
\end{algorithm}

\textbf{Claims.}
All experiments compare StocBiO~\citep{ji2021bilevel} and SUSTAIN~\citep{khanduri2021near} as stochastic bilevel baselines against our ANE methods. In particular, we compare SUSTAIN against ANE-VR, since both methods include variance reduction. We verify two claims: \emph{(C1) The Neumann product exhibits nonzero odd-order perturbative contributions whenever the per-factor Hessian noise has nonzero third central moment $\mu_3'$. Although standardized scalar skewness scales as $m^{-1/2}$ in the leading perturbative approximation, the scale-sensitive third-moment diagnostic $\mathcal M_3(\xi)=\E[\|\xi-\E\xi\|^3]$ can grow with truncation depth $m$ and conditioning $\kappa_g$} (Theorem~\ref{thm:skew_scalar}, Proposition~\ref{prop:vector_third}). {Lemma~\ref{lem:sum_growth} proves this growth for the multiplicative model under weak contraction.} \emph{(C2) Antithetic cancellation reduces the absolute third-moment proxy vs. {a single} standard Neumann {estimate}; this is associated with improved predictive metrics on the reported tasks.}

\textbf{Synthetic.}
We consider a synthetic bilevel problem with inner quadratic $g(\blam,\bbeta)=\tfrac12\bbeta^\top H\bbeta-\blam^\top C\bbeta$, prescribed $\kappa_g$, and stochastic HVPs $\hat H v=s_i\cdot Hv$ with $\E[s_i]=1$ and positive third central moment $\E[(s_i-1)^3]>0$ (right-skewed). Sweeping $\kappa_g\in\{10,100,1000\}$ and $m\in\{1,\ldots,200\}$ at $d=50$, Figure~\ref{fig:teaser} is consistent with the mechanism isolated by Theorem~\ref{thm:skew_scalar} and Proposition~\ref{prop:odd_cancel}: under asymmetric scalar Hessian noise and weak contraction, the absolute third-moment proxy is lower for ANE than for {a single} standard Neumann {estimate} ($41$--$74\%$ reduction across the swept $\kappa_g$); the theorem explains the odd perturbation-degree terms while the plotted proxy is an empirical absolute third-moment diagnostic.
\paragraph{{Matched controls.}}
{Matched synthetic controls ($\kappa_g=100$, $m=20$, eight seeds) isolate the asymmetry effect. With distributional mirroring, in which the mirrored branch draws its own multiplier, ANE {matches Standard~(2x), which averages two estimates, in variance and stochastic-HVP budget, yet has} a much smaller coordinate skewness ($0.42\pm0.43$ versus $1.30\pm0.13$); reference-only gives $1.87\pm0.19$. In a fixed-variance sweep the two are close for symmetric multipliers and separate as the multipliers become skewed (Appendix~\ref{app:matched_controls}).}

\begin{figure*}[!htbp]
\centering
\begin{subfigure}[t]{0.49\textwidth}
  \centering
  \includegraphics[width=\textwidth]{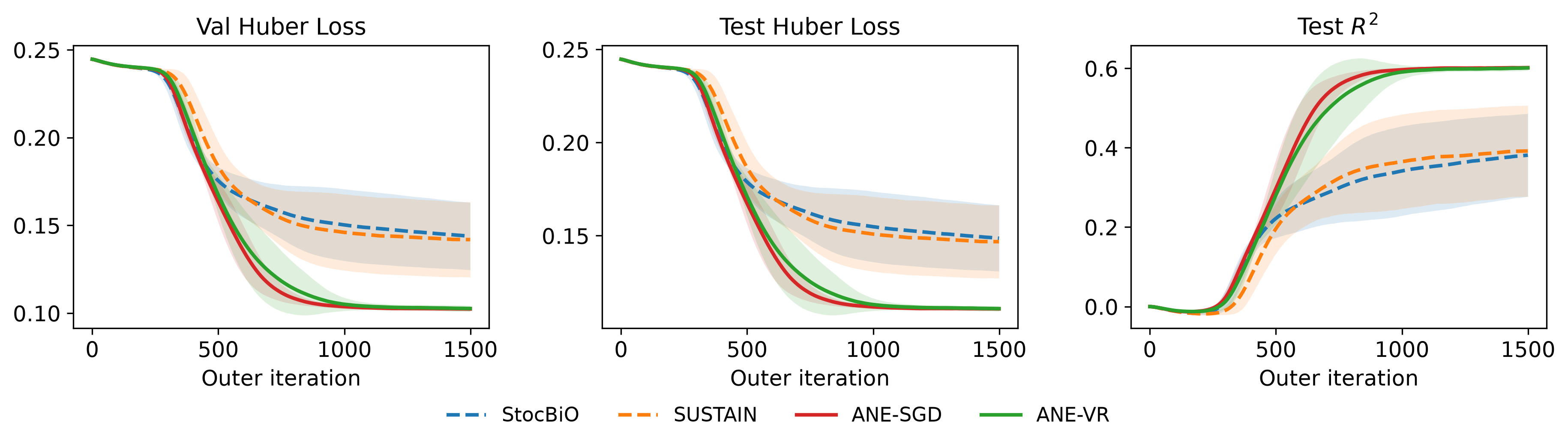}
  \caption{California Housing.}
  \label{fig:task1_conv}
\end{subfigure}\hfill
\begin{subfigure}[t]{0.49\textwidth}
  \centering
  \includegraphics[width=\textwidth]{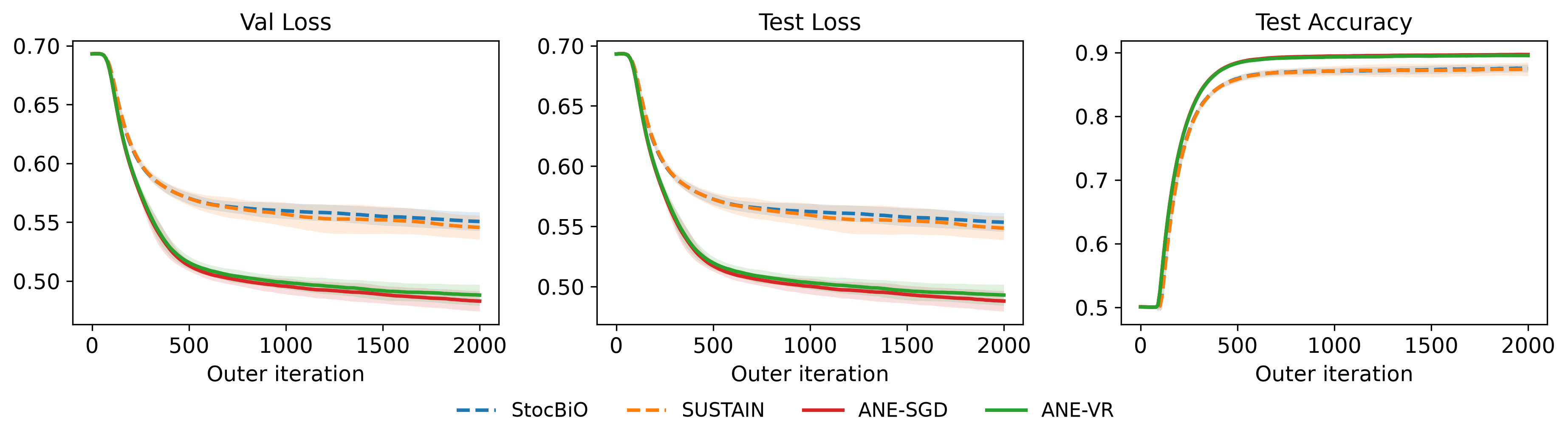}
  \caption{{Fashion-MNIST}.}
  \label{fig:task2_conv}
\end{subfigure}

\vspace{0.6em}

\begin{subfigure}[t]{0.49\textwidth}
  \centering
  \includegraphics[width=\textwidth]{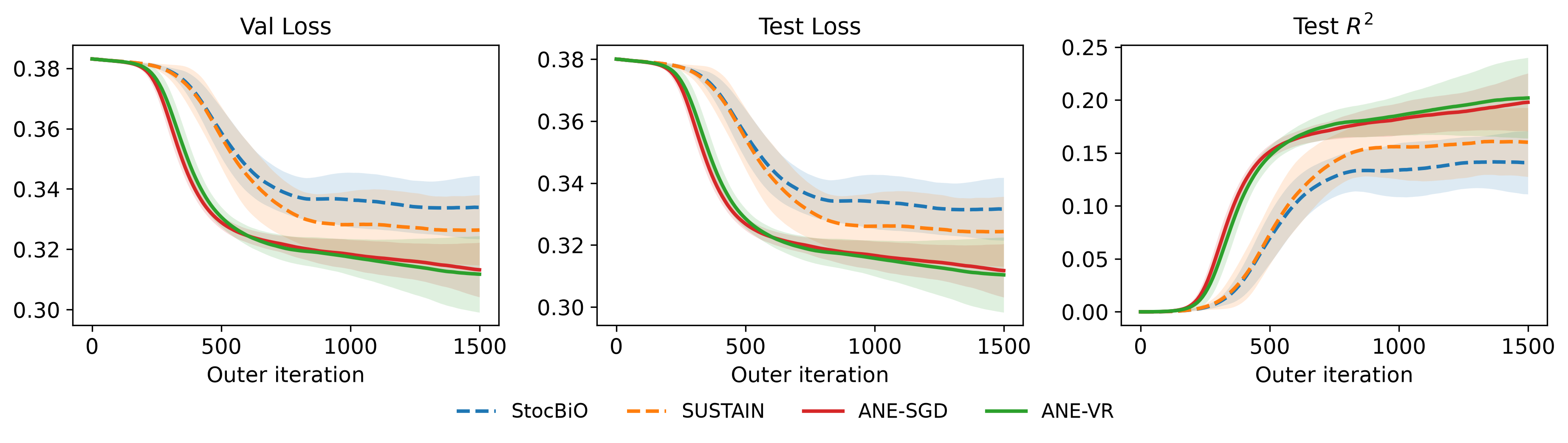}
  \caption{Acrobot.}
  \label{fig:task3a_conv}
\end{subfigure}\hfill
\begin{subfigure}[t]{0.49\textwidth}
  \centering
  \includegraphics[width=\textwidth]{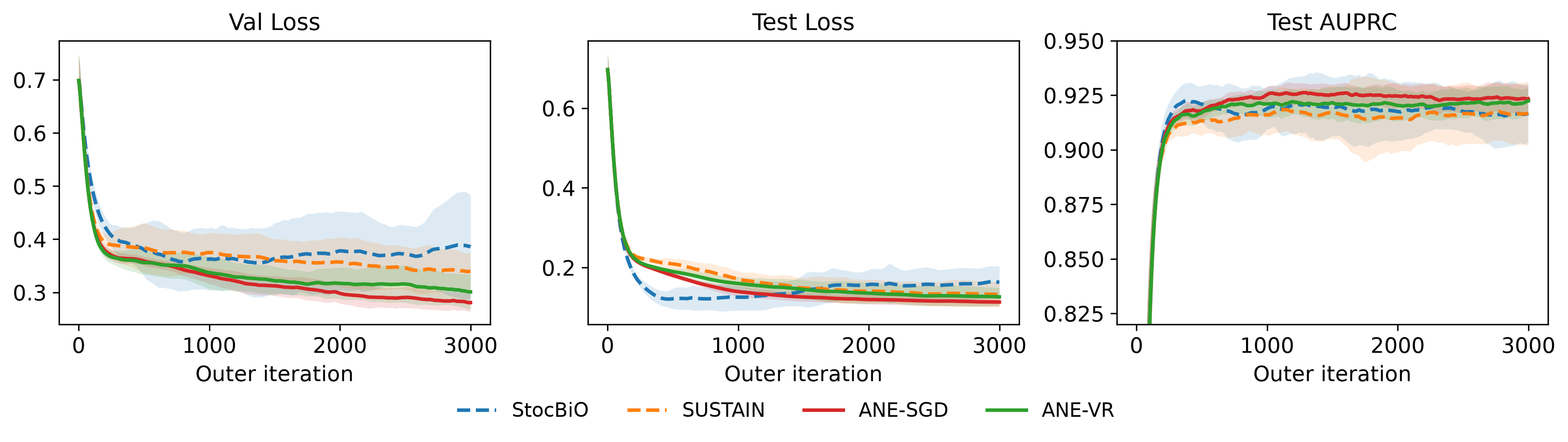}
  \caption{Fraud detection.}
  \label{fig:task4_conv}
\end{subfigure}
\caption{\textbf{Convergence on bilevel tasks.} Each panel: validation (outer) loss, test loss, and the task-specific test metric (mean $\pm$ {SD over 20 seeds}) vs. outer iteration. Dashed: baselines (StocBiO, SUSTAIN); solid: ANE. {On Housing, Fashion, and Acrobot,} ANE-SGD and ANE-VR separate from baselines early and maintain a consistent gap{; on Fraud, StocBiO has the lowest test loss mid-training and the AUPRC bands overlap}. {Appendix~\ref{app:trajectories} enlarges each panel{; curves are smoothed with an exponential moving average}.}}
\label{fig:real_tasks}
\end{figure*}

\textbf{Real tasks.}
We evaluate ANE-SGD/ANE-VR on four bilevel problems (Table~\ref{tab:tasks}); Table~\ref{tab:results} reports final metrics and Figure~\ref{fig:real_tasks} convergence trajectories (mean $\pm$ {SD over 20 seeds}). $\sigma_i''$ is per-sample loss curvature: when $\sigma_i''\approx0$ for some samples and $\gg0$ for others, HVP factors $(I-\sigma_i''J_i^\top J_i/\hat\eta)$ {have asymmetric noise} and odd-degree contributions accumulate across Neumann factors (Theorem~\ref{thm:skew_scalar}; per-task derivations and hyperparameters in Appendix~\ref{app:exp_details}).

\begin{table}[t]
\centering
\caption{\textbf{Summary of bilevel tasks.} The asymmetry source column identifies the origin of nonzero per-factor $\mu_3'$ in the Neumann product for each task. Appendix~\ref{app:per_task_analysis} {motivates} this for each task.}
\label{tab:tasks}
\small
\begin{tabular}{@{}lllp{3.6cm}@{}}
\toprule
Task & Problem & Dataset & Asymmetry source \\
\midrule
Housing & Robust regr.       & {California Housing}~\citep{pace1997sparse} & 20\% corrupted targets \\
Fashion & Data cleaning      & {Fashion-MNIST}~\citep{xiao2017fashion} & \leavevmode{10\% of positives flipped} \\
Acrobot & Offline RL         & Acrobot-v1~\citep{brockman2016openai} & Bimodal episode lengths \\
Fraud   & Imbalanced classif.& creditcard~\citep{dal2015calibrating} & 6\% minority class \\
\bottomrule
\end{tabular}
\end{table}

\textbf{Task 1 (Housing).}\label{sec:exp_housing}
Pseudo-Huber regression {on California Housing}~\citep{pace1997sparse} with 20\% corrupted targets; outer learns per-feature regularization on clean validation. Corrupted targets give $\sigma_i''\approx0$ for ${\sim}20\%$ of samples (worse conditioning, amplified odd-degree contributions, Theorem~\ref{thm:skew_scalar}). ANE reaches $R^2=0.60$ vs. $0.39$ ($+54\%$); loss $0.147\to0.110$ ($-25\%$, Figure~\ref{fig:task1_conv}).

\textbf{Task 2 ({Fashion-MNIST}).}
MLP/softplus with {one-directional} label corruption {(10\% of positive labels flipped)}~\citep{xiao2017fashion}; outer tunes {per-feature regularization} on clean validation. $\sigma_i''\approx0.25$ for corrupted vs. $\approx0$ for clean (asymmetric noise). ANE-SGD/ANE-VR reach {89.6\%/89.5\% vs. 87.5\%/87.6\%} (${\sim}2${\,}pp), test loss {$-11\%$} (Figure~\ref{fig:task2_conv}).

\textbf{Task 3 (Acrobot).}
MLP value function on Acrobot-v1~\citep{brockman2016openai}; offline dataset 50\% expert (${\sim}136$ steps) / 50\% random (${\sim}500$ steps). Random episodes are ${\sim}80\%$ of transitions but $\sigma_i''\approx0$, leaving ${\sim}21\%$ effective curvature. ANE-SGD/ANE-VR reach $R^2=0.20$ vs. $0.16$ (SUSTAIN), $0.14$ (StocBiO); loss {$0.311$ vs.\ $0.325$ (SUSTAIN; $-4\%$}, Figure~\ref{fig:task3a_conv}).

\textbf{Task 4 (Fraud).}
MLP on credit-card transactions~\citep{dal2015calibrating}; outer uses class-balanced validation. Boundary fraud ($\sigma_i''\approx0.25$) vs. high-margin legitimate ($\sigma_i''\approx0$): strong noise asymmetry. ANE-SGD/ANE-VR cut outer loss to {$0.281/0.301$ vs. $0.339$ (SUSTAIN; 11--17\%), {$+0.5$--$0.6$\,pp} AUPRC {(vs.\ StocBiO)}} (Figure~\ref{fig:task4_conv}).

\begin{table}[t]
\centering
\scriptsize
\setlength{\tabcolsep}{3pt}
\caption{\textbf{Final test metrics (mean $\pm$ {SD over 20 seeds per task}).} {For Fraud, Loss is the class-balanced validation (outer) objective (Appendix~\ref{app:hyperparams}).} Best per task in \textbf{bold}. In Acrobot, SUSTAIN beats StocBiO ($R^2=0.16$ vs. $0.14$) but ANE-VR matches ANE-SGD{, consistent with across-iteration VR adding little on top of ANE}.}\label{tab:results}
\begin{tabular}{@{}llcccc@{}}
\toprule
Task & Metric & StocBiO & SUSTAIN & ANE-SGD & ANE-VR \\
\midrule
Housing & Loss $\downarrow$ & {$.147\pm.018$} & {$.147\pm.020$} & {$\mathbf{.110\pm.001}$} & {$\mathbf{.111\pm.001}$} \\
 & $R^2\uparrow$ & {$.388\pm.104$} & {$.393\pm.116$} & {$\mathbf{.604\pm.007}$} & {$\mathbf{.602\pm.007}$} \\
\addlinespace
Fashion & Loss $\downarrow$ & {$.553\pm.008$} & {$.548\pm.011$} & {$\mathbf{.487\pm.009}$} & {$\mathbf{.492\pm.009}$} \\
 & Acc $\uparrow$ & {$87.5\pm0.8$} & {$87.6\pm1.3$} & {$\mathbf{89.6\pm0.3}$} & {$\mathbf{89.5\pm0.3}$} \\
\addlinespace
Acrobot & Loss $\downarrow$ & {$.332\pm.011$} & {$.325\pm.012$} & {$\mathbf{.311\pm.010}$} & {$\mathbf{.310\pm.013}$} \\
 & $R^2\uparrow$ & {$.140\pm.031$} & {$.159\pm.034$} & {$\mathbf{.200\pm.032}$} & {$\mathbf{.203\pm.040}$} \\
\addlinespace
Fraud & Loss $\downarrow$ & {$.380\pm.089$} & {$.339\pm.035$} & {$\mathbf{.281\pm.017}$} & {$\mathbf{.301\pm.033}$} \\
 & AUPRC $\uparrow$ & {$.918\pm.012$} & {$.916\pm.017$} & {$\mathbf{.924\pm.006}$} & {$\mathbf{.923\pm.009}$} \\
\bottomrule
\end{tabular}
\end{table}

\textbf{Ablation and cost.}
An ablation over $\tau\in\{0.5,0.7,0.9,1.0\}$ (Appendix~\ref{app:tau_ablation}) shows {that, with 20 seeds per value, all four mirror strengths give nearly identical predictive metrics (differences of at most $0.004$ in loss, $R^2$, or AUPRC and $0.1$\,pp in accuracy), all well separated from the baselines.}
{The hypergradient-step sweep in Appendix~\ref{app:timing_sweep}, measured with each task's configuration, reports $2.7$--$3.3\times$ ANE-SGD overhead across $m=5,10,20,50$.}

\section{Conclusion and Limitations}\label{sec:conclusion}

{The product structure of the Neumann series introduces a third source of error beyond bias and variance, skewness: the third-moment error grows with truncation depth and inner condition number under weak contraction (Lemma~\ref{lem:sum_growth}), and with per-factor noise asymmetry (Theorem~\ref{thm:skew_scalar}). ANE pairs each Hessian sample with a mirrored counterpart around a reference Hessian. With an exact reference and $\tau=1$ only even-order noise remains, so at fixed depth the inverse-Hessian third-moment term in the iterate-level bound (Theorem~\ref{thm:iterate}) is {$O(\rho^6)$ in the Hessian-error level $\rho<1$, versus $O(\rho^3)$} for Standard; with an approximate one, under uniform contraction, the odd-order residual {is bounded linearly in} the reference error. ANE modifies only the Hessian--vector step; ANE-SGD and ANE-VR lower loss by {4}--25\% relative to the best stochastic or variance-reduced baseline on every task (20 seeds). Skewness reduction is thus a {promising}, complementary mechanism for improving hypergradient quality.}
\emph{Limitations:}\label{sec:limitations} odd-order cancellation {is} not a full hypergradient skewness theory{; the approximate-reference bound assumes a contraction the real tasks lack (Appendix~\ref{app:real_diagnostics}), so the gains are not attributable to skewness correction alone. ANE costs about three times StocBiO's hypergradient step (Appendix~\ref{app:timing_sweep}); matched-wall-clock evaluation is future work.}

{\renewcommand{\acksection}{\section*{{Acknowledgments}}}
\begin{ack}
We sincerely thank Bloomberg's CTO Office for its support of the Visiting Faculty Program, through which Pawe{\l} Polak participated in this research.
\end{ack}
}

\newpage
\bibliographystyle{abbrvnat}
  \bibliography{references}

\newpage
\appendix

 \textbf{Appendix roadmap.}
Appendix~\ref{app:skew_proofs}   {gives estimator proofs, the full-sum lower bounds, the conditional-moment iterate bound, and small-noise moment scaling. Appendix~\ref{app:algorithms} records detailed algorithms}. Appendix~\ref{app:exp_details} provides   {hyperparameters, controls, runtime measurements, and diagnostic tables. Appendix~\ref{app:trajectories} shows the 20-seed convergence trajectories.
}

\section{Skewness Analysis Proofs}\label{app:skew_proofs}

\subsection{Proof of Theorem~\ref{thm:skew_scalar}}\label{app:scalar_proof}
\begin{proof}
Let $X_j=a+\delta_j$ with $\delta_j$ i.i.d., $\E[\delta_j]=0$, $\E[\delta_j^2]=\sigma^2$, and $\E[\delta_j^3]=\mu_3'$. Define
\[
  P_m\triangleq\prod_{j=1}^m X_j.
\]
\textbf{Part (i).} By independence,
\begin{equation}
  \E[P_m]=\prod_{j=1}^m\E[X_j]=a^m.
\end{equation}
Also,
\begin{equation}
  \E[P_m^2]=\prod_{j=1}^m\E[X_j^2]=(a^2+\sigma^2)^m,
\end{equation}
so
\begin{equation}
  \Var(P_m)=(a^2+\sigma^2)^m-a^{2m}.
\end{equation}
\textbf{Part (ii).} Since
\[
  \E[X_j^3]=a^3+3a\sigma^2+\mu_3',
\]
we have
\begin{equation}
  \E[P_m^3]=(a^3+3a\sigma^2+\mu_3')^m.
\end{equation}
Using the identity $\E[(Y-\E Y)^3]=\E[Y^3]-3\E[Y^2]\E[Y]+2(\E[Y])^3$ with $Y=P_m$ yields
\begin{align}
  \E[(P_m-a^m)^3]
  &=\E[P_m^3]-3\E[P_m^2]\E[P_m]+2(\E[P_m])^3\\
  &=(a^3+3a\sigma^2+\mu_3')^m-3(a^2+\sigma^2)^m a^m+2a^{3m},
\end{align}
which is~\eqref{eq:mu3_product}.

\textbf{Part (iii).} For the perturbative expansion we write $r\triangleq\sigma^2/a^2$ and $s\triangleq\mu_3'/a^3$. If the Neumann scale is chosen so that all scalar factors are positive, then $a\in(0,1)$. More generally, the following moment expansions hold for $0<|a|<1$; the denominator in the standardized skewness uses $|a|$, which introduces the factor $\operatorname{sgn}(a)^{m-1}$. Then
\begin{equation}
  \Var(P_m)=a^{2m}\bigl[(1+r)^m-1\bigr].
\end{equation}
Using $(1+r)^m=1+mr+\binom{m}{2}r^2+O(m^3r^3)$ we get
\[
  \Var(P_m)=m\sigma^2a^{2m-2}+O(m^2\sigma^4|a|^{2m-4}),
\]
which is the variance expansion of Theorem~\ref{thm:skew_scalar}(iii). Here $m\sigma^2/a^2=o(1)$ implies $m^3r^3=o(m^2r^2)$.

For the third central moment,~\eqref{eq:mu3_product} rewrites as $\E[(P_m-a^m)^3]=a^{3m}[(1+3r+s)^m-3(1+r)^m+2]$ and expanding gives
\[
 (1+3r+s)^m-3(1+r)^m+2
 =ms+\binom{m}{2}(6r^2+6rs+s^2)+O\bigl(m^3(r^3+r^2|s|+rs^2+|s|^3)\bigr).
\]
Therefore
\[
  \E[(P_m-a^m)^3]
  =m\mu_3'a^{3m-3}+O(m^2\sigma^4|a|^{3m-4})+O(m^2\sigma^2|\mu_3'||a|^{3m-5})+O(m^2\mu_3'^2|a|^{3m-6}).
\]
Under fixed $a$ bounded away from zero and sufficiently small $\sigma$, with $|\mu_3'|\le C\sigma^3$, the last two error terms scale as $O(m^2\sigma^5|a|^{3m-5})$ and $O(m^2\sigma^6|a|^{3m-6})$ respectively and are dominated by $O(m^2\sigma^4|a|^{3m-4})$. Making the absolute value explicit when expanding the denominator,
\[
  \Var(P_m)^{3/2}=m^{3/2}\sigma^3|a|^{3m-3}\bigl(1+O(m\sigma^2/a^2)\bigr),
\]
so
\[
  \gamma_1(P_m)=\frac{m\mu_3'a^{3m-3}}{m^{3/2}\sigma^3|a|^{3m-3}}+O(\sqrt m\,\sigma)=\operatorname{sgn}(a)^{m-1}\frac{\mu_3'}{\sigma^3\sqrt m}+O(\sqrt m\,\sigma),
\]
which is the standardized skewness expansion of Theorem~\ref{thm:skew_scalar}(iii).
\end{proof}

\subsection{Explicit next-order skewness expansion}\label{app:scalar_expansion}

\begin{remark}[Explicit next-order standardized-skewness expansion]\label{rem:explicit_next_order}
With $r \triangleq \sigma^2/a^2$ and $s \triangleq \mu_3'/a^3$, the variance and third central moment of $P_m$ admit the explicit binomial expansions
\begin{equation}\label{eq:scalar_explicit_den}
  \frac{\operatorname{Var}(P_m)}{a^{2m}} = mr + \binom{m}{2}r^2 + O(m^3 r^3),
\end{equation}
\begin{equation}\label{eq:scalar_explicit_num}
  \frac{\E[(P_m-a^m)^3]}{a^{3m}} = ms + \binom{m}{2}(6r^2 + 6rs + s^2) + O\bigl(m^3(r^3 + r^2|s| + r s^2 + |s|^3)\bigr).
\end{equation}
Consequently, {for $a>0$,} the standardized skewness expands as
\begin{equation}\label{eq:scalar_explicit_gamma}
  \gamma_1(P_m)
  = \frac{\mu_3'}{\sigma^3\sqrt m}
  + O\!\left(\sqrt m\,\frac{\sigma}{|a|}\right)
  + O\!\left(\sqrt m\,\frac{|\mu_3'|}{a^2\sigma}\right).
\end{equation}
In the bounded-shape perturbative regime where $|a|$ is bounded away from zero, $|\mu_3'| \le C\sigma^3$, and $m\sigma^2/a^2 = o(1)$, the last two error terms collapse to $O(\sqrt m\,\sigma)$ and \eqref{eq:scalar_explicit_gamma} reduces to the form stated in Theorem~\ref{thm:skew_scalar}(iii).
\end{remark}

\begin{proof}
With \(r = \sigma^2/a^2\) and \(s = \mu_3'/a^3\), the exact identities from Theorem~\ref{thm:skew_scalar} become
\begin{align}
  \frac{\operatorname{Var}(P_m)}{a^{2m}} &= (1+r)^m - 1, \\
  \frac{\mathbb E[(P_m-a^m)^3]}{a^{3m}} &= (1+3r+s)^m - 3(1+r)^m + 2.
\end{align}
Applying the binomial expansion to the variance term yields
\begin{equation}
  \frac{\operatorname{Var}(P_m)}{a^{2m}}
  = mr + \binom{m}{2}r^2 + O(m^3r^3),
\end{equation}
which is \eqref{eq:scalar_explicit_den}.
For the numerator,
\begin{align}
  (1+3r+s)^m
  &= 1 + m(3r+s) + \binom{m}{2}(3r+s)^2 \\
  &\quad + O\bigl(m^3(|r|+|s|)^3\bigr),
\end{align}
and therefore
\begin{align}
  \frac{\mathbb E[(P_m-a^m)^3]}{a^{3m}}
  &= \Bigl[1 + m(3r+s) + \binom{m}{2}(3r+s)^2\Bigr] \\
  &\quad - 3\Bigl[1 + mr + \binom{m}{2}r^2\Bigr] + 2 \\
  &\quad + O\bigl(m^3(r^3 + r^2|s| + r s^2 + |s|^3)\bigr) \\
  &= ms + \binom{m}{2}(6r^2 + 6rs + s^2) \\
  &\quad + O\bigl(m^3(r^3 + r^2|s| + r s^2 + |s|^3)\bigr),
\end{align}
which is \eqref{eq:scalar_explicit_num}.
Using
\[
  (mr + \binom{m}{2}r^2 + O(m^3r^3))^{3/2}
  = (mr)^{3/2}\bigl(1 + O(mr)\bigr),
\]
we obtain
\begin{align}
  \gamma_1(P_m)
  &= \frac{ms + O\bigl(m^2(r^2 + r|s| + s^2)\bigr)}{(mr)^{3/2}\bigl(1+O(mr)\bigr)} \\
  &= \frac{s}{r^{3/2}\sqrt m}
  + O\!\left(\frac{\sqrt m\,r^{1/2}}{1}\right)
  + O\!\left(\frac{\sqrt m\,|s|}{r^{1/2}}\right).
\end{align}
Returning to \((\sigma,\mu_3',a)\) {with \(a>0\)} gives
\begin{equation}
  \gamma_1(P_m)
  = \frac{\mu_3'}{\sigma^3\sqrt m}
  + O\!\left(\sqrt m\,\frac{\sigma}{|a|}\right)
  + O\!\left(\sqrt m\,\frac{|\mu_3'|}{a^2\sigma}\right),
\end{equation}
which is \eqref{eq:scalar_explicit_gamma}. If \(|a|\) is bounded away from zero and
\(|\mu_3'| \le C\sigma^3\), the last two terms are both \(O(\sqrt m\,\sigma)\).
\end{proof}

\subsection{Proof of Proposition~\ref{prop:odd_cancel} and Corollary~\ref{cor:quadratic_bound}}\label{app:odd_cancel}
\begin{proof}[Proof of Proposition~\ref{prop:odd_cancel}]
Fix a depth \(k\). By distributivity,
\begin{equation}
  P_k^{\pm}
  = \prod_{j=1}^k (A \pm E_j)
  = \sum_{S \subseteq [k]} (\pm 1)^{|S|}\,\Pi_k(S),
\end{equation}
where \(\Pi_k(S)\) is the ordered product with factor \(E_j\) when \(j \in S\) and factor
\(A\) otherwise. Averaging the two expansions gives
\begin{equation}
  \frac{P_k^+ + P_k^-}{2}
  = \sum_{S \subseteq [k]} \frac{1+(-1)^{|S|}}{2}\,\Pi_k(S)
  = \sum_{\substack{S \subseteq [k] \\ |S|\ \mathrm{even}}} \Pi_k(S),
\end{equation}
which is exactly \eqref{eq:even_order_expansion}.
\end{proof}

\begin{proof}[Proof of Corollary~\ref{cor:quadratic_bound}]
For each depth \(k\), Proposition~\ref{prop:odd_cancel} shows that
\(\tfrac12(P_k^+ + P_k^-) - A^k\) contains only even-order perturbation terms. Every ordered product
with exactly \(\ell\) perturbation factors has norm at most \(a^{k-\ell}\rho^{\ell}\), and there are
\(\binom{k}{\ell}\) ways to choose the locations of those perturbation factors. Therefore,
\begin{equation}\label{eq:pk_even_bound_app}
  \left\|\frac{P_k^+ + P_k^-}{2} - A^k\right\|
  \le
  \sum_{\substack{\ell=2 \\ \ell\ \mathrm{even}}}^{k}
  \binom{k}{\ell} a^{k-\ell}\rho^{\ell}.
\end{equation}
Using the elementary identity
\begin{equation}
  \sum_{\substack{\ell=0 \\ \ell\ \mathrm{even}}}^{k}
  \binom{k}{\ell} a^{k-\ell}\rho^{\ell}
  = \frac{(a+\rho)^k + (a-\rho)^k}{2},
\end{equation}
we obtain
\begin{equation}
  \left\|\frac{P_k^+ + P_k^-}{2} - A^k\right\|
  \le
  \frac{(a+\rho)^k + (a-\rho)^k - 2a^k}{2}.
\end{equation}
Summing over \(k=0,\ldots,m-1\) and multiplying by \(\|b\|/\hat\eta\) yields \eqref{eq:quadratic_bound}.

For the asymptotic form, expand the right-hand side in \(\rho\) around \(0\):
\[
  (a+\rho)^k + (a-\rho)^k - 2a^k
  = k(k-1)a^{k-2}\rho^2 + O(\rho^4).
\]
For fixed \(a\le a_0<1\),
\[
  \sum_{k=0}^{\infty} k(k-1)a^{k-2}
  = \frac{2}{(1-a)^3}
  \le \frac{2}{(1-a_0)^3},
\]
so the quadratic term is uniformly
\[
  O\!\left(\frac{\rho^2\|b\|}{\hat\eta(1-a_0)^3}\right).
\]
If no gap from one is imposed, then using \(a^{k-2}\le 1\) gives
\(\sum_{k=0}^{m-1}k(k-1)=O(m^3)\), so the finite-depth weak-contraction bound is
\(O(m^3\rho^2\|b\|/\hat\eta)\). This proves the corollary.
\end{proof}

\subsection{Proof of Proposition~\ref{prop:reference_error}}\label{app:reference_error}
\begin{proof}
Only the mirrored branch changes when \(H\) is replaced by \(\bar H\). For a fixed depth \(k\), define
\begin{equation}
  M_j \triangleq A + E_j,
  \qquad
  \widetilde M_j \triangleq A - B + E_j.
\end{equation}
Then the noncommutative telescoping identity gives
\begin{equation}
  \prod_{j=1}^{k}\widetilde M_j - \prod_{j=1}^{k} M_j
  =
  \sum_{r=1}^{k}
  \left(\prod_{j=1}^{r-1}\widetilde M_j\right)
  (\widetilde M_r - M_r)
  \left(\prod_{j=r+1}^{k} M_j\right).
\end{equation}
Since \(\widetilde M_r - M_r = -B\) and each factor norm is at most \(c\), we obtain
\begin{equation}
  \left\|\prod_{j=1}^{k}\widetilde M_j - \prod_{j=1}^{k} M_j\right\|
  \le k c^{k-1}\|B\|.
\end{equation}
Summing over \(k=1,\ldots,m-1\) and including the prefactor \(1/(2\hat\eta)\) from the average of the two
branches yields
\begin{equation}
  \bigl\|\hat v_{\rm ANE}(\bar H) - \hat v_{\rm ANE}(H)\bigr\|
  \le
  \frac{\|B\|\,\|b\|}{2\hat\eta}
  \sum_{k=1}^{m-1} k c^{k-1}.
\end{equation}
Finally, because \(c<1\),
\begin{equation}
  \sum_{k=1}^{m-1} k c^{k-1}
  \le
  \sum_{k=1}^{\infty} k c^{k-1}
  = \frac{1}{(1-c)^2},
\end{equation}
which proves \eqref{eq:reference_error_bound}.
\end{proof}

\subsection{Proof of Proposition~\ref{prop:skew_budget}}\label{app:skew_budget_proof}
\begin{proof}
Write
\[
  G = \nabla_{\blam} f - J v^\star,
  \qquad
  \widehat G = \widehat\nabla_{\blam} f - \widehat J\,\hat v.
\]
Add and subtract \(J\hat v\):
\begin{align}
  \widehat G - G
  &= \bigl(\widehat\nabla_{\blam} f - \nabla_{\blam} f\bigr)
     - \widehat J\,\hat v + J v^\star \\
  &= \bigl(\widehat\nabla_{\blam} f - \nabla_{\blam} f\bigr)
     - (\widehat J - J)\hat v - J(\hat v - v^\star),
\end{align}
which is~\eqref{eq:full_hg_error_decomp}.

For the cubic norm bound, define
\[
  x \triangleq \widehat\nabla_{\blam} f - \nabla_{\blam} f,
  \qquad
  y \triangleq -(\widehat J - J)\hat v,
  \qquad
  z \triangleq -J(\hat v - v^\star).
\]
Then
\[
  \|\widehat G - G\|^3 = \|x+y+z\|^3 \le (\|x\|+\|y\|+\|z\|)^3.
\]
For nonnegative numbers \(a,b,c\), Hölder's inequality implies
\[
  (a+b+c)^3 \le 3^2(a^3+b^3+c^3) = 9(a^3+b^3+c^3).
\]
Applying this with \(a=\|x\|\), \(b=\|y\|\), and \(c=\|z\|\) yields
\begin{equation}
  \|\widehat G - G\|^3
  \le
  9\|x\|^3 + 9\|y\|^3 + 9\|z\|^3,
\end{equation}
that is,
\begin{equation}
  \|\widehat G - G\|^3
  \leq
  9\bigl\|\widehat\nabla_{\blam} f - \nabla_{\blam} f\bigr\|^3
  + 9\bigl\|(\widehat J-J)\hat v\bigr\|^3
  + 9\|J\|^3\,\bigl\|\hat v - v^\star\bigr\|^3,
\end{equation}
which is~\eqref{eq:full_hg_third_bound}. The final sentence of the proposition follows because the explicit
truncation-depth parameter \(m\) appears only through the inverse-Hessian{--}vector estimator \(\hat v\).
\end{proof}

\begingroup
\subsection{\texorpdfstring{{Lower bounds for the full truncated sum}}{Lower bounds for the full truncated sum}}
\label{app:sum_growth}

{Claim~(C1) in Section~\ref{sec:experiments} states that the third-moment proxy $\mathcal M_3$ can grow with truncation depth $m$ and conditioning $\kappa_g$; this subsection proves it for the full truncated Neumann sum in the multiplicative model under weak contraction. The growth is compatible with the $m^{-1/2}$ decay of standardized skewness in Theorem~\ref{thm:skew_scalar}, which concerns a single product.
}

\begin{lemma}[Full-sum lower bounds]
\label{lem:sum_growth}
{Let $0<a<1$ and let $\delta_1,\ldots,\delta_{m-1}$ be i.i.d.\ with $\E\delta_j=0$, $\E\delta_j^2=\sigma^2$, $\E|\delta_j|^3<\infty$, and $\nu_3\triangleq\E\delta_j^3$ {(the $\mu_3'$ of Theorem~\ref{thm:skew_scalar})}. Let $P_k=\prod_{j=1}^k(a+\delta_j)$ and $S_m=\sum_{k=1}^{m-1}(P_k-a^k)$. For $m\geq2$,
}\begin{equation}\label{eq:sum_second_lower}
\E|S_m|^3\geq\left(\E S_m^2\right)^{3/2}\geq\sigma^3a^{3m-6}\left[\frac{(m-1)m(2m-1)}{6}\right]^{3/2},
\end{equation}
{and, if $\nu_3\geq0$,
}\begin{equation}\label{eq:sum_third_lower}
\E|S_m|^3\geq\E S_m^3\geq\nu_3\sum_{k,l,n=1}^{m-1}a^{k+l+n-3}\min(k,l,n)\geq\nu_3a^{3m-6}\left[\frac{m(m-1)}{2}\right]^2.
\end{equation}
{If a coordinate of the Neumann error satisfies $\xi_i-\E\xi_i=(b_i/\hat\eta)S_m$, then $\mathcal M_3(\xi)\geq(|b_i|/\hat\eta)^3\E|S_m|^3$. If also $a\geq1/2$ and $m(1-a)\leq C$, the two bounds are $\Omega(\sigma^3m^{9/2})$ and $\Omega(\nu_3m^4)$, with constants depending only on $C$.
}\end{lemma}

\begin{proof}
{The first inequality in \eqref{eq:sum_second_lower} is Lyapunov's. For $k\leq l$, $P_l=P_k\prod_{j=k+1}^{l}(a+\delta_j)$, where the product is independent of $P_k$ and has mean $a^{l-k}$, so $\Cov(P_k,P_l)=a^{l-k}[(a^2+\sigma^2)^k-a^{2k}]\geq k\sigma^2a^{k+l-2}$. Summing over $k,l$,
}\[
\E S_m^2\geq\sigma^2a^{2m-4}\sum_{k,l=1}^{m-1}\min(k,l)=\sigma^2a^{2m-4}\,\frac{(m-1)m(2m-1)}{6}.
\]
{For \eqref{eq:sum_third_lower}, $P_k-a^k=\sum_{\emptyset\neq U\subseteq\{1,\ldots,k\}}a^{k-|U|}\prod_{j\in U}\delta_j$ has nonnegative coefficients. In $\E S_m^3$, monomials in which some index appears once vanish, and every other index contributes $\sigma^2$ or $\nu_3\geq0$, so all terms are nonnegative. Keeping, for each $j\leq\min(k,l,n)$, the product of the three terms linear in $\delta_j$ gives $\nu_3a^{k+l+n-3}\min(k,l,n)$, and $\sum_{k,l,n}\min(k,l,n)=\sum_{j=1}^{m-1}(m-j)^3=[m(m-1)/2]^2$. The vector bound follows from $\|\xi-\E\xi\|\geq|\xi_i-\E\xi_i|$, and the rates from $a^{3m-6}\geq e^{-6C}$, since $\log a\geq-2(1-a)$ for $a\geq1/2$.
}\end{proof}

\paragraph{{Sign and conditioning.}}
{In the multiplicative model $\delta_j=-(\mu_i/\hat\eta)(s_j-1)$, so $\nu_3=-(\mu_i/\hat\eta)^3\E[(s_j-1)^3]$: right-skewed multipliers, as in Figure~\ref{fig:teaser}, give $\nu_3<0$, and only \eqref{eq:sum_second_lower} applies. Along the smallest eigenvalue, with $b_i$ and $\hat\eta\propto\ell_{g,1}$ fixed, $1-a=\Theta(1/\kappa_g)$ and $\sigma\propto\kappa_g^{-1}$; choosing $m=\Theta(\kappa_g)$ keeps $m(1-a)$ bounded, so the bound in \eqref{eq:sum_second_lower} grows as $\kappa_g^{3/2}$ and, when $\nu_3\geq0$, the bound in \eqref{eq:sum_third_lower} grows as $\kappa_g$.
}
\endgroup

\begingroup
\subsection{\texorpdfstring{{An iterate-level bound with a third-order remainder}}{An iterate-level bound with a third-order remainder}}
\label{app:outer_descent}

{This subsection proves the curvature-weighted bound of Theorem~\ref{thm:iterate} as Proposition~\ref{prop:outer_descent}, which also records the first-order special case. The proposition carries the conditional bias, curvature-weighted variance, and centered third moment of the update direction $d_t$ (the ANE-SGD estimate, or the recursive ANE-VR direction provided its carried error satisfies the bias bound) through to a stationarity bound on the outer iterates. The proof expands $F$ to second order: the third moment enters through the cubic Taylor remainder, and keeping the exact curvature gives the curvature-weighted variance.
}

\begin{proposition}[Conditional-moment stationarity bound]
\label{prop:outer_descent}
{Let $F:\R^p\to\R$ be bounded below by $F_\star$, with $L_F$-Lipschitz gradient and $L_2$-Lipschitz Hessian, and let $\blam_{t+1}=\blam_t-\alpha d_t$ with $0<\alpha\leq1/(4L_F)$. Let $\mathcal F_t$ contain $\blam_t$ and all randomness preceding $d_t$; write $r_t=\E[d_t\mid\mathcal F_t]-\nabla F(\blam_t)$, $e_t=d_t-\E[d_t\mid\mathcal F_t]$, and $y_t=\blam_t-\alpha\E[d_t\mid\mathcal F_t]$; and suppose that, almost surely and uniformly in $t$, $\|r_t\|^2\leq\bar r^2$, $\E[\|e_t\|^2\mid\mathcal F_t]\leq\bar\sigma^2$, $\E[\|e_t\|^3\mid\mathcal F_t]\leq\bar M_3$, and $\E[e_t^\top\nabla^2F(y_t)e_t\mid\mathcal F_t]\leq\bar Q$. Then
}\begin{equation}\label{eq:iterate_bound_curv}
\frac{1}T\sum_{t=0}^{T-1}\E\|\nabla F(\blam_t)\|^2
\leq
\frac{2(F(\blam_0)-F_\star)}{\alpha T}
+\frac{5}{2}\bar r^2+\alpha\bar Q
+\frac{L_2\alpha^2}{3}\bar M_3.
\end{equation}
{Since $\nabla^2F\preceq L_FI$, one may take $\bar Q=L_F\bar\sigma^2$, which gives the first-order bound:
}\begin{equation}\label{eq:iterate_bound}
\frac{1}T\sum_{t=0}^{T-1}\E\|\nabla F(\blam_t)\|^2
\leq
\frac{2(F(\blam_0)-F_\star)}{\alpha T}
+\frac{5}{2}\bar r^2+L_F\alpha\bar\sigma^2
+\frac{L_2\alpha^2}{3}\bar M_3.
\end{equation}
\end{proposition}

\begin{proof}
{Condition on $\mathcal F_t$ and abbreviate $x=\blam_t$, $g=\nabla F(x)$, $h=g+r_t$, and $y=y_t=x-\alpha h$. By $L_F$-smoothness, $-\langle g,r_t\rangle\leq\|g\|^2/4+\|r_t\|^2$, $\|g+r_t\|^2\leq2\|g\|^2+2\|r_t\|^2$, and $L_F\alpha\leq1/4$,
}\[
F(y)\leq F(x)-\alpha\langle g,h\rangle+\tfrac12L_F\alpha^2\|h\|^2\leq F(x)-\tfrac\alpha2\|g\|^2+\tfrac54\alpha\|r_t\|^2.
\]
{The second-order Taylor bound about $y$ gives
}\[
F(y-\alpha e_t)\leq F(y)-\alpha\langle\nabla F(y),e_t\rangle+\frac{\alpha^2}{2}e_t^\top\nabla^2F(y)e_t+\frac{L_2\alpha^3}{6}\|e_t\|^3.
\]
{Since $y$ is $\mathcal F_t$-measurable, the linear term has zero conditional mean and the quadratic term has conditional mean at most $\tfrac{\alpha^2}{2}\bar Q$. Combining the two inequalities, taking expectations, summing over $t$, dividing by $\alpha T/2$, and using $\E F(\blam_T)\geq F_\star$ proves \eqref{eq:iterate_bound_curv}.
}\end{proof}

\subsection{\texorpdfstring{{Hessian-error level and centered third moments}}{Hessian-error level and centered third moments}}
\label{app:moment_scaling}

{This subsection bounds the inverse-Hessian component of the third-moment term $\bar M_3$ in Theorem~\ref{thm:iterate}. {For independent factor perturbations at Hessian-error level $\rho\le1$ (Assumption~\ref{ass:higher}(i)), with an exact reference and $\tau=1$, ANE's centered third moment is $O(\rho^6)$ at every depth, versus $O(\rho^3)$ for the standard Neumann estimator; the constants are uniform in $m$ under a contraction margin.}
}

\begin{proposition}[{{Third-moment scaling in the Hessian-error level}}]
\label{prop:moment_scaling}
{Fix $m$, $A$, $b$, and $\hat\eta$ with $\|A\|\leq a<1$, and assume an exact reference and $\tau=1$. {Let the scaled errors $E_j=(\hat H_j-H)/\hat\eta$ be independent with $\E E_j=0$ and $(\E\|E_j\|^3)^{1/3}\leq\rho\le1$. Then}
}\[
{\E\|\hat v_{\rm Neu}-\mathcal N_m(H;b)\|^3\le\Bigl(\frac{K_1\|b\|}{\hat\eta}\Bigr)^{3}\rho^3,
\qquad
\E\|\hat v_{\rm ANE}-\mathcal N_m(H;b)\|^3\le\Bigl(\frac{K_2\|b\|}{\hat\eta}\Bigr)^{3}\rho^6,}
\]
{{with $K_1=\sum_{k=1}^{m-1}k(1+a)^{k-1}$ and $K_2=\sum_{k=2}^{m-1}\binom{k}{2}(1+a)^{k-2}$; if moreover $a+\rho\le c<1$, then $K_1$ and $K_2$ can be replaced by $(1-c)^{-2}$ and $(1-c)^{-3}$, uniformly in $m$.} Both estimators have mean $\mathcal N_m(H;b)$, so these are their centered third moments.
}\end{proposition}

\begin{proof}
{For an ordered monomial with $\ell$ noise factors, submultiplicativity and
independence give its $L^3$ norm at most
$a^{k-\ell}{\rho^\ell}$.
Minkowski's inequality applied to the ordered expansions gives
}\begin{align*}
\left(\E\|\hat v_{\rm Neu}-\mathcal N_m\|^3\right)^{1/3}
&\leq\frac{\|b\|}{\hat\eta}
\sum_{k=0}^{m-1}\sum_{\ell=1}^k
\binom{k}{\ell}a^{k-\ell}{\rho^\ell},\\
\left(\E\|\hat v_{\rm ANE}-\mathcal N_m\|^3\right)^{1/3}
&\leq\frac{\|b\|}{\hat\eta}
\sum_{k=0}^{m-1}\sum_{\substack{\ell=2\\\ell\ {\rm even}}}^k
\binom{k}{\ell}a^{k-\ell}{\rho^\ell}.
\end{align*}
{{Since $\rho\le1$, $(a+\rho)^k-a^k\le k\rho(1+a)^{k-1}$ bounds the first sum by $\rho K_1$. In the second, $\binom{k}{\ell}\le\binom{k}{2}\binom{k-2}{\ell-2}$ for $\ell\ge2$ bounds each depth by $\rho^2\binom{k}{2}(a+\rho)^{k-2}\le\rho^2\binom{k}{2}(1+a)^{k-2}$, hence the sum by $\rho^2K_2$; with $a+\rho\le c<1$ the same steps give $\rho/(1-c)^2$ and $\rho^2/(1-c)^3$. The second sum keeps only the even-$\ell$ terms of the first, so it is no larger for every $\rho$. Including the $\ell=0$ terms, the same expansion gives $(\E\|\hat v\|^3)^{1/3}\le\frac{\|b\|}{\hat\eta}\sum_{k=0}^{m-1}(a+\rho)^k$ for both estimators, so the Jacobian term of $\bar M_3$, which involves $\hat v$, is bounded alike.} Every nonconstant ordered monomial {contains each $E_j$ at most once, so it} has
zero expectation by independence and centering, proving the mean identity.
}\end{proof}
\endgroup

\clearpage
\section{\texorpdfstring{{Detailed Algorithms}}{Detailed Algorithms}}\label{app:algorithms}
\begin{algorithm}[H]
\caption{ANE-SGD: antithetic Neumann hypergradients without recursive variance reduction}
\label{alg:ane_sgd}
\begin{flushleft}
\textbf{Input:} outer steps $T$, inner steps $K$, truncation depth $m$, outer step size $\alpha$, inner step size $\eta$, Neumann scale $\hat\eta$, mirror strength $\tau \in [0,1]$, reference-refresh period $q_{\mathrm{ref}}$.\\[3pt]
\textbf{Initialize:} $\blam_0$, $\bbeta_0$, reference Hessian $\bar H_0$.\\[3pt]
\textbf{for} $t=0,1,\ldots,T-1$ \textbf{do}
\begin{enumerate}
  \item \textbf{Inner solve.}
  Starting from $\bbeta_t$, run $K$ stochastic-gradient steps on $g(\blam_t,\cdot)$ with step size $\eta$ to obtain $\bbeta_{t+1}$.

  \item \textbf{Reference refresh.}
  If $t \equiv 0 \pmod{q_{\mathrm{ref}}}$, update $\bar H_t$ using a large-batch or full-batch Hessian evaluation at $(\blam_t,\bbeta_{t+1})$; otherwise reuse the previous reference (or update it by an exponential moving average). {The experiments use a reference batch four times the HVP batch, applied as an HVP operator.}

  \item \textbf{ANE hypergradient.}
  Draw stochastic Hessians $\widehat H_{t,1},\ldots,\widehat H_{t,m-1}$. For each $j$, set
  \[
    \widehat H_{t,j}^+ = \widehat H_{t,j},
    \qquad
    \widehat H_{t,j}^- = \bar H_t - \tau(\widehat H_{t,j}-\bar H_t).
  \]
  Form
  \[
    \hat v_{\rm ANE}
    =
    \frac{1}{2\hat\eta}
    \sum_{k=0}^{m-1}
    \Biggl[
      \prod_{j=1}^{k}\Bigl(I-\widehat H_{t,j}^+/\hat\eta\Bigr)
      +
      \prod_{j=1}^{k}\Bigl(I-\widehat H_{t,j}^-/\hat\eta\Bigr)
    \Biggr] b_t,
  \]
  where $b_t = \widehat\nabla_{\bbeta} f(\blam_t,\bbeta_{t+1})$. Then set
  \[
    \widehat G_t
    \leftarrow
    \widehat\nabla_{\blam} f(\blam_t,\bbeta_{t+1})
    -
    \widehat J_t\,\hat v_{\rm ANE},
    \qquad
    \widehat J_t = \widehat\nabla^2_{\blam\bbeta} g(\blam_t,\bbeta_{t+1}).
  \]

  \item \textbf{Outer step.}
  Update $\blam_{t+1} \leftarrow \blam_t - \alpha\widehat G_t$ (or its projected/proximal variant).
\end{enumerate}
\textbf{end for}
\end{flushleft}
\end{algorithm}

\begin{algorithm}[H]
\caption{ANE-VR: antithetic Neumann hypergradients with recursive outer variance reduction}
\label{alg:ane_vr}
\begin{flushleft}
\textbf{Input:} outer steps $T$, inner steps $K$, truncation depth $m$, outer step size $\alpha$, inner step size $\eta$, Neumann scale $\hat\eta$, mirror strength $\tau \in [0,1]$, reference-refresh period $q_{\mathrm{ref}}$, {snapshot probability $p$}.\\[3pt]
\textbf{Initialize:} $\blam_0$, $\bbeta_0$, reference Hessian $\bar H_0$, direction $d_{-1}=0$.\\[3pt]
\textbf{for} $t=0,1,\ldots,T-1$ \textbf{do}
\begin{enumerate}
  \item \textbf{Inner solve.}
  Starting from $\bbeta_t$, run $K$ stochastic-gradient steps on $g(\blam_t,\cdot)$ to obtain $\bbeta_{t+1}$.

  \item \textbf{Reference refresh.}
  If $t \equiv 0 \pmod{q_{\mathrm{ref}}}$, refresh $\bar H_t$ using a large batch or a full batch at $(\blam_t,\bbeta_{t+1})$; otherwise reuse or exponentially average the previous reference.

  \item \textbf{ANE hypergradient.}
  Form $\widehat G_t^{\rm ANE}$ exactly as in Algorithm~\ref{alg:ane_sgd}, but with the mini-batch choice used by the outer variance-reduction wrapper.

  \item \textbf{Recursive variance reduction.}
  {At $t=0$ and otherwise with probability $p$, set $d_t \leftarrow \widehat G_t^{\rm ANE}$. Otherwise, set $d_t \leftarrow d_{t-1} + \widehat G_t^{\rm ANE} - \widehat G^{\rm ANE}(\blam_{t-1},\bbeta_{t-1})$.}

  \item \textbf{Outer step.}
  Update $\blam_{t+1} \leftarrow \blam_t - \alpha d_t$ (or its projected/proximal variant).
\end{enumerate}
\textbf{end for}
\end{flushleft}
\end{algorithm}

\clearpage
\section{Extended Evaluation Details}\label{app:exp_details}

This appendix records full experimental details for Section~\ref{sec:experiments}.

\subsection{Estimator Diagnostics}

Given $L$ independent hypergradient estimates $\{\widehat G^{(\ell)}\}_{\ell=1}^L$ with mean $\bar G$, the primary estimator diagnostic is the empirical third-moment error:
  \begin{equation}\label{eq:third_moment_diag}
  {\widehat{\mathcal M}_3(\widehat G-G)}
  =
  \frac{1}{L}\sum_{\ell=1}^L \|\widehat G^{(\ell)} - \bar G\|^3.
\end{equation}
This is a conservative third-moment proxy that   upper-bounds directional   {signed third central moments}: $|\E[(u^\top(\xi-\E\xi))^3]| \leq \E[\|\xi-\E\xi\|^3]$ for any unit $u$.   {It measures centered fluctuation scale and need not vanish for symmetric errors. Signed third moments and standardized skewness require a fixed projection $u$: $Y^{(\ell)}=u^\top\widehat G^{(\ell)}$, $\hat\mu_3=L^{-1}\sum_\ell(Y^{(\ell)}-\bar Y)^3$, and $\hat\gamma_1=\hat\mu_3/[L^{-1}\sum_\ell(Y^{(\ell)}-\bar Y)^2]^{3/2}$. Bias requires a reference $G$, and vector variance is $L^{-1}\sum_\ell\|\widehat G^{(\ell)}-\bar G\|^2$. The raw error moment $L^{-1}\sum_\ell\|\widehat G^{(\ell)}-G\|^3$ is distinct from the centered proxy.}

\subsection{Synthetic Benchmark}\label{app:synthetic_benchmark}
The inner objective is $g(\blam,\bbeta) = \tfrac{1}{2}\bbeta^\top H\bbeta - \blam^\top C\bbeta$ with $H = \mathrm{diag}(\mu_g, \ldots, \ell_{g,1})$ having prescribed condition number $\kappa_g = \ell_{g,1}/\mu_g$, {with $\ell_{g,1}$} fixed at $L_{\max}=10$ {and} $\mu_g = L_{\max}/\kappa_g$. The coupling matrix $C$ is random orthogonal. Stochastic HVPs use $s_i \cdot Hv$ where $s_i$ are i.i.d.\ with $\E[s_i]=1$ and positive third central moment. Dimension $d=50$, 5 seeds, $L=1000$ estimator samples per configuration. We sweep $\kappa_g \in \{10,100,1000\}$ and $m \in \{1,\ldots,200\}$. {In Figure~\ref{fig:teaser} the mirrored branch draws its own multiplier (distributional mirroring; Section~\ref{sec:experiments}).} {The real tasks mirror the sampled mini-batch Hessian itself, as in~\eqref{eq:mirror_hessian}.}

\subsection{Real Task Hyperparameters}\label{app:hyperparams}

\begin{table}[H]
\centering
\caption{Full hyperparameter settings for each real task.}
\label{tab:hyperparams}
\begin{tabular}{@{}lcccc@{}}
\toprule
Parameter & Task 1 & Task 2 & Task 3 & Task 4 \\
 & (Housing) & (Fashion) & (Acrobot) & (Fraud) \\
\midrule
$K$ (inner steps) & 10 & 50 & 10 & 10 \\
$\eta$ (inner lr) & 0.01 & 0.01 & 0.01 & 0.01 \\
$\alpha$ (outer lr) & 0.01 & 0.001 & 0.01 & 0.01 \\
Outer optimizer & Adam & Adam & Adam & Adam \\
$m$ (truncation) & 50 & 20 & 50 & 10 \\
$\hat\eta$ & {$\hat L_{\max}/2$} & {$\hat L_{\max}/2$} & {$\hat L_{\max}/2$} & {$\hat L_{\max}/2$} \\
Batch size ($g,f$) & {8} & {32} & {8} & 8 \\
Batch size ($H$) & 4 & 8 & 4 & 4 \\
$T$ (outer iters) & 1500 & 2000 & 1500 & 3000 \\
Seeds & 20 & {20} & 20 & {20} \\
{Gradient clip} & {1000} & {100} & {1000} & {1000} \\
\bottomrule
\end{tabular}
\end{table}

{ANE uses $\tau=1$ in all tasks, with a safeguard that damps $\tau$ for an individual factor only when the mirrored step would be unstable (the damped mirroring of Remark~\ref{rem:stabilization}). {ANE also caps the mirrored perturbation $\|\hat Hv-\bar Hv\|$ at $5\|v\|$, and both ANE and the standard Neumann estimator rescale a branch state whose norm exceeds $100\|b\|$ (the branch-norm guard). $\hat L_{\max}$ is a power-iteration estimate of the largest eigenvalue of the full-training inner Hessian near initialization.} The reference is an HVP operator over a batch four times the HVP batch, drawn at the refresh point and refreshed every $q_{\rm ref}=10$ steps; a dense Hessian is never formed. Adam is the outer optimizer in these experiments, so Proposition~\ref{prop:outer_descent} is a theoretical result for a separate constant-step gradient update.}

\paragraph{{Models and data.}}
{Housing uses a pseudo-Huber loss with $\delta=0.5$ and an MLP with 32 hidden units on 8 PCA components of the inputs. We corrupt $20\%$ of the training targets via $y\mapsto -y+3\,\mathrm{sign}(y)$; validation {and test} targets are clean. Fashion separates classes 0--4 (positive) from 5--9 with a softplus loss, per-feature regularization, and an MLP with 16 hidden units on 50 PCA components. It has $10{,}000$ training samples, with $10\%$ of their positive labels flipped to negative. Acrobot fits an MLP $9\to32\to1$ to return-to-go targets from 500 episodes, half expert and half random, using a pseudo-Huber loss with $\delta=1$. Fraud trains an MLP $30\to32\to1$ on $4{,}500$ legitimate and $295$ fraud transactions with per-parameter regularization. Its outer objective, a class-balanced validation loss, is also the reported loss.} {The data come from sklearn (California Housing), torchvision (Fashion-MNIST), Kaggle (credit-card fraud), and Gymnasium (Acrobot-v1).} {Housing and Acrobot split the samples (transitions) at random into $60/20/20\%$ train/validation/test{, so Acrobot transitions from one episode can fall in several splits}; Fashion holds out $2{,}500$ validation samples and uses the official test set; Fraud splits the fraud cases $60/20/20\%$ and adds $1{,}000$ legitimate validation transactions, with a test set of the same class balance. The Acrobot expert is an energy-pumping heuristic policy and the random policy samples actions uniformly.}

\subsection{Mirror Strength Ablation}\label{app:tau_ablation}

Table~\ref{tab:tau_ablation} ablates $\tau \in \{0.5, 0.7, 0.9, 1.0\}$,   where $\tau=1$ is the   {a priori default. Smaller} $\tau$   {damps the mirror perturbation; at zero it averages Standard with a deterministic reference branch. All columns use 20 seeds (mean $\pm$ SD). The predictive metrics vary little with $\tau$ relative to the ANE--baseline gaps; on Fraud, $\tau=1$ attains both the lowest loss and the highest AUPRC.}

\begin{table}[H]
\centering
\small
 \setlength{\tabcolsep}{5pt}
 \caption{  \textbf{ANE-SGD sensitivity to mirror strength $\tau$} (mean $\pm$   {SD over 20 seeds for every column}). SUSTAIN  {is} shown for reference.}
\label{tab:tau_ablation}

\input{tables/tau_ablation.tex}

\end{table}

\begingroup
\subsection{\texorpdfstring{{Controlled Estimator Comparisons}}{Controlled Estimator Comparisons}}\label{app:matched_controls}

{We distinguish estimator diagnostics from downstream predictive metrics. In the synthetic controls, skewness is summarized by the coordinate statistic $|\gamma_1|$, the mean over coordinates of the absolute standardized skewness of the estimate; it removes the scale of each coordinate's variance and differs from the centered absolute vector moment $\mathcal M_3({\xi})=\E\|{\xi}-\E{\xi}\|^3$, which depends on scale and need not vanish under symmetric noise. A reduction in either diagnostic alone does not identify the cause of every downstream improvement.
}

{All controls use the synthetic problem with $\kappa_g=100$, $m=20$, multiplier standard deviation $0.3$, $L=2000$ estimates per run, and eight seeds (mean $\pm$ SD over seeds). {The multipliers are $s=1+0.3Z$, with $Z$ a standardized gamma variable of the target skewness (a clipped Gaussian at skewness $0$). Tables~\ref{tab:matched_controls}--\ref{tab:tau_skewness} share their Standard draws but draw ANE's mirrored multipliers at different points of the random stream, so their ANE entries at the shared setting (multiplier skewness $4$, $\tau=1$) agree only within seed variability.} Standard~(2x) averages two independent standard Neumann estimates and so has ANE's stochastic-HVP budget. Reference-only is the $\tau=0$ construction, which uses the reference without mirroring.
}

\begin{table}[H]
\centering
\small
\caption{{Matched controls with distributional mirroring (multiplier skewness $4$).}}
\label{tab:matched_controls}
\input{tables/controls_matched_legacy.tex}
\end{table}

{At matched stochastic-HVP budget and variance, ANE's coordinate skewness is far below that of Standard~(2x), while reference-only keeps the skewness of Standard at a quarter of its variance (with a fixed exact reference, the $\tau=0$ estimate is half of Standard plus a constant, so its standardized skewness equals Standard's by construction).
}

\begin{table}[H]
\centering
\small
\caption{{Fixed-variance sweep with distributional mirroring: the multiplier variance is fixed and its standardized skewness is varied.}}
\label{tab:fixed_variance_sweep}
\input{tables/controls_sweep_legacy.tex}
\end{table}

{In Table~\ref{tab:fixed_variance_sweep}, the standard control's skewness increases with the skewness of the multipliers, while ANE's remains substantially lower at the three nonzero targets, at the same variance. For symmetric multipliers the two are close: this negative control shows no detectable advantage in the measured diagnostic.}
\endgroup

\begingroup
\subsection{\texorpdfstring{{Mirror Strength and Distributional Asymmetry}}{Mirror Strength and Distributional Asymmetry}}\label{app:tau_skewness}

{For an exact reference, a perturbation term of odd degree $n$ has coefficient $(1-\tau^n)/2$ in the averaged construction, relative to the Standard coefficient. Thus $\tau=0.5$ retains one quarter of the leading linear coefficient, whereas $\tau=1$ cancels it exactly. Table~\ref{tab:tau_skewness} records the skewness sweep over $\tau$ with distributional mirroring (same setting as Appendix~\ref{app:matched_controls}, multiplier skewness $4$), together with $\mathcal M_3$ and the variance at each mirror strength.
}

\begin{table}[H]
\centering
\small
\caption{{Mirror-strength sweep with distributional mirroring (mean $\pm$ SD over eight seeds).}}
\label{tab:tau_skewness}
\resizebox{\ifdim\width>\linewidth\linewidth\else\width\fi}{!}{\input{tables/controls_tau_legacy.tex}}
\end{table}

{The skewness diagnostic continues to improve beyond $\tau=0.5$, even when the task metrics in Table~\ref{tab:tau_ablation} change only modestly, and $\mathcal M_3$ stays $69$--$88\%$ below Standard at every mirror strength. Because $\mathcal M_3$ is scale-sensitive and also reflects variance, which rises toward $\tau=1$ in this construction, its behavior need not coincide with the skewness curve. We use $\tau=1$ as the a priori default for exact odd-degree cancellation.}
\endgroup

\begingroup
\subsection{\texorpdfstring{{Real-Task Reference and Estimator Diagnostics}}{Real-Task Reference and Estimator Diagnostics}}\label{app:real_diagnostics}\label{app:reference_diagnostics}

{This section reports, for the real tasks, the error of ANE's training reference, whether the Neumann recursion is contractive as Proposition~\ref{prop:reference_error} assumes, and, for completeness, the per-task estimator moments. All diagnostics use each task's training configuration (Table~\ref{tab:hyperparams}) at checkpoints $t$ of a seed-0 ANE-SGD run of $T$ outer steps, with that run's $\hat\eta$, mirror strength, safeguard, and current reference $\bar H$.
}

\paragraph{{Reference error and contraction.}}
{ANE's reference (an HVP operator on a batch four times the HVP batch, refreshed every ten steps) has relative error $0.06$--$1.99$ across tasks and checkpoints (Table~\ref{tab:reference_error}). At the training $\hat\eta=\hat L_{\max}/2$, the depth-$m$ products are not contractive: their per-factor rates $(\|\prod_{j=1}^m\hat A_j^\pm b\|/\|b\|)^{1/m}$, averaged over eight draws, exceed one at most checkpoints ($0.92$--$3.03$ for the forward and $1.30$--$9.26$ for the raw $\tau=1$ mirrored branch), and on Acrobot the products overflow at several checkpoints. Proposition~\ref{prop:reference_error} therefore does not apply on these tasks, and both estimators stay bounded only through the branch-norm guard and, for ANE, the $\tau$ safeguard.
}

\begin{table}[H]
\centering
\small
\caption{{Relative error $\|\bar H-H\|/\|H\|$ of ANE's training reference $\bar H$ with respect to the full training Hessian $H$ (spectral norms by power iteration) at five checkpoints of the seed-0 ANE-SGD run.}}
\label{tab:reference_error}
\input{tables/diag_reference_error.tex}
\end{table}

\paragraph{{Estimator moments.}}
{Table~\ref{tab:real_estimator_diagnostics} lists the per-task bias, variance, and skewness of Standard, Standard~(2x), and ANE at each task's training depth; because the recursion is not contractive at the training $\hat\eta$, these magnitudes reflect the safeguards rather than antithetic cancellation.
}

\begin{table}[H]
\centering
\small
\caption{{Estimator moments at each task's training depth $m$ and mid-training checkpoint $t=T/2$ (mean $\pm$ SD over four repetitions of $L=250$ estimates $\hat g$). Bias is $\|\E\hat g-g_{\rm ref}\|$ and variance $\E\|\hat g-\E\hat g\|^2$, with $g_{\rm ref}$ a deep Standard estimate (depth $200$, batch $512$, $32$ draws) at the training $\hat\eta$; $\gamma_1$ is the standardized skewness of $u^\top\hat g$ along the fixed direction $u=g_{\rm ref}/\|g_{\rm ref}\|$; coordinate $|\gamma_1|$ is the mean over coordinates of the absolute standardized skewness. Standard~(2x) averages two Standard estimates (ANE's stochastic-HVP budget).}}
\label{tab:real_estimator_diagnostics}
\scriptsize
\setlength{\tabcolsep}{4pt}
\resizebox{\ifdim\width>\linewidth\linewidth\else\width\fi}{!}{\input{tables/diag_estimators.tex}}
\end{table}
\endgroup

\begingroup
\subsection{\texorpdfstring{{Runtime Across Truncation Depths}}{Runtime Across Truncation Depths}}\label{app:timing_sweep}

{Table~\ref{tab:timing_sweep} reports the time of one hypergradient computation at a shared mid-training checkpoint for $m\in\{5,10,20,50\}$, with each task's configuration (Table~\ref{tab:hyperparams}, which also lists the task depths). Times exclude the shared $K$-step inner solve and are means over three seeds of $20$ calls on one CPU thread (Apple M4 Pro). They include ANE's reference handling: the refresh is negligible, and applying the reference costs about one stochastic HVP per mirrored factor.
}

\begin{table}[H]
\centering
\small
\caption{{Hypergradient-step wall-clock time (ms per call; ratio to StocBiO in parentheses).}}
\label{tab:timing_sweep}
\input{tables/timing.tex}
\end{table}

{Across tasks and depths, ANE-SGD costs $2.7$--$3.3$ times StocBiO's hypergradient step ($3.0$--$3.1$ at the task depths), consistent with three HVP-type operations per factor against one; ANE-VR costs $3.1$--$4.0$ times and SUSTAIN $1.5$--$1.7$ times, reflecting the extra computation in their PAGE steps.
}

{The mirrored Hessian acts as $\hat H^-v=(1+\tau)\bar H v-\tau\hat H v$. It can be applied through a reference-batch HVP without storing a dense $p\times p$ Hessian. Each factor remains an HVP computation with dependence on the model and reference-batch size, while ANE adds another branch and reference applications.
}
\endgroup

\subsection{Per-Task Asymmetry Analysis}\label{app:per_task_analysis}

{An illustrative source of curvature asymmetry is the generalized Gauss--Newton term $b^{-1}\sum_{i\in B}\sigma_i''J_i^\top J_i$. For a nonlinear model, the full loss Hessian also contains $\ell'(z_i)\nabla^2 z_i$ (and regularization); the displayed term is therefore not an identity for every MLP Hessian. If directional Jacobian factors are treated as fixed, mini-batch factor noise has third moment proportional to
}\begin{equation}\label{eq:mu3_link}
 \mu_3'\propto-\frac{1}{b^2}\E[(\sigma_i''-\bar\sigma'')^3].
\end{equation}
  {For a two-point curvature model with active fraction $p$ and active curvature $c$, the exact identity is
}\begin{equation}\label{eq:bernoulli_mu3}
 \E[(\sigma_i''-\bar\sigma'')^3]=pc^3(1-p)(1-2p).
\end{equation}
{This is nonzero away from $p\in\{0,1/2,1\}$. Its absolute value is not monotone as $p$ moves toward zero; the standardized two-point skewness is $(1-2p)/\sqrt{p(1-p)}$. The following task descriptions motivate asymmetry qualitatively rather than certify scalar-model assumptions for the full Hessian.
}

\paragraph{Ill-conditioning from reduced effective samples ($\kappa_g$).}
  {For a small threshold $\varepsilon_{\rm curv}>0$, let $n_{\rm eff}=|\{i:\sigma_i''\ge\varepsilon_{\rm curv}\}|$. Curvature concentrated on a small subset can reduce directional coverage in the Gauss--Newton term and worsen conditioning. Pseudo-Huber curvature is strictly positive at finite residuals, so this counts substantial curvature rather than literal nonzero curvature.}

In \emph{Task~1 (Housing)}, {pseudo-Huber} regression gives $\sigma_i'' = (1 + (r_i/\delta)^2)^{-3/2}$ where $r_i = y_i - f(x_i;\bbeta)$ is the residual.
Corrupted targets (20\%) produce large residuals with $\sigma_i'' \approx 0$, while clean targets (80\%) yield $\sigma_i'' > 0$.
This reduces $n_{\mathrm{eff}}$ to ${\sim}80\%$; the MLP Jacobian $J_i$ varies with hidden-layer activations, so the reduced active subset provides non-uniform directional coverage,   {which can worsen conditioning.}
 The resulting bimodal curvature distribution ($p=0.8$ clean, $0.2$ corrupted) gives $\E[(\sigma_i''-\bar\sigma'')^3]\neq 0$ via~\eqref{eq:bernoulli_mu3}, so $\mu_3'\neq 0$.

In \emph{Task~3 (Acrobot)}, the same {pseudo-Huber} loss applies with $\sigma_i'' = (1 + (r_i/\delta)^2)^{-3/2}$.
Expert episodes (${\sim}136$ steps) and random episodes (${\sim}500$ steps) contribute unequal transition counts: random transitions outnumber expert ones ${\sim}4{:}1$ and have large residuals with $\sigma_i'' \approx 0$, reducing $n_{\mathrm{eff}}$ to ${\sim}21\%$ of the training data.
The bimodal curvature distribution ($p\approx 0.21$ expert, $0.79$ random) gives $\mu_3'\neq 0$ via~{\eqref{eq:mu3_link}--\eqref{eq:bernoulli_mu3}},   {motivating possible conditioning and asymmetry effects in this approximation.}

\paragraph{Noise asymmetry from bimodal curvature ($\mu_3'$).}

In \emph{Task~2 (Fashion)}, softplus classification gives $\sigma_i'' = \mathrm{sigmoid}(z_i)(1 - \mathrm{sigmoid}(z_i))$ where $z_i = y_i f(x_i;\bbeta)$ is the margin.
Corrupted samples resist confident classification ($|z_i|$ small, $\sigma_i'' \approx 0.25$); clean samples are classified confidently ($|z_i|$ large, $\sigma_i'' \approx 0$).
With {$p\approx0.05$ (10\% of the positive half)}, equation~\eqref{eq:bernoulli_mu3} gives $\E[(\sigma_i''-\bar\sigma'')^3]>0$, so $\mu_3'\neq 0$.

In \emph{Task~4 (Fraud)}, the same softplus loss applies.
Fraud samples ($p = 0.06$) lie near the decision boundary ($\sigma_i'' \approx 0.25$); legitimate samples have large margins ($\sigma_i'' \approx 0$).
  {The minority fraction can produce large standardized curvature skewness; it does not imply the largest raw third moment across tasks, since curvature scales and directional Jacobian factors differ.}

 \paragraph{{Interpretation}.}
  {These mixture arguments explain why the per-sample curvature can be asymmetric, but they do not show that the scalar perturbative regime holds on a given task. At the mid-training and final checkpoints, the training Hessian is ill-conditioned on Housing ($\hat\kappa_g\approx240$--$340$) and near-singular on Acrobot; on Fashion it is near-singular at mid-training and reaches $\hat\kappa_g\approx180$ by the end. By contrast, Fraud is well-conditioned ($\hat\kappa_g\approx40$) and has by far the largest per-factor curvature skewness.
}

\section{\texorpdfstring{{Enlarged Convergence Trajectories}}{Enlarged Convergence Trajectories}}\label{app:trajectories}
{These are enlarged versions of the panels in Figure~\ref{fig:real_tasks} (mean $\pm$ SD over 20 seeds). Curves are smoothed with an exponential moving average ({smoothing factor} $0.1$) before averaging.}

\begin{figure}[H]
\centering
\begin{subfigure}{\textwidth}
\includegraphics[width=\textwidth]{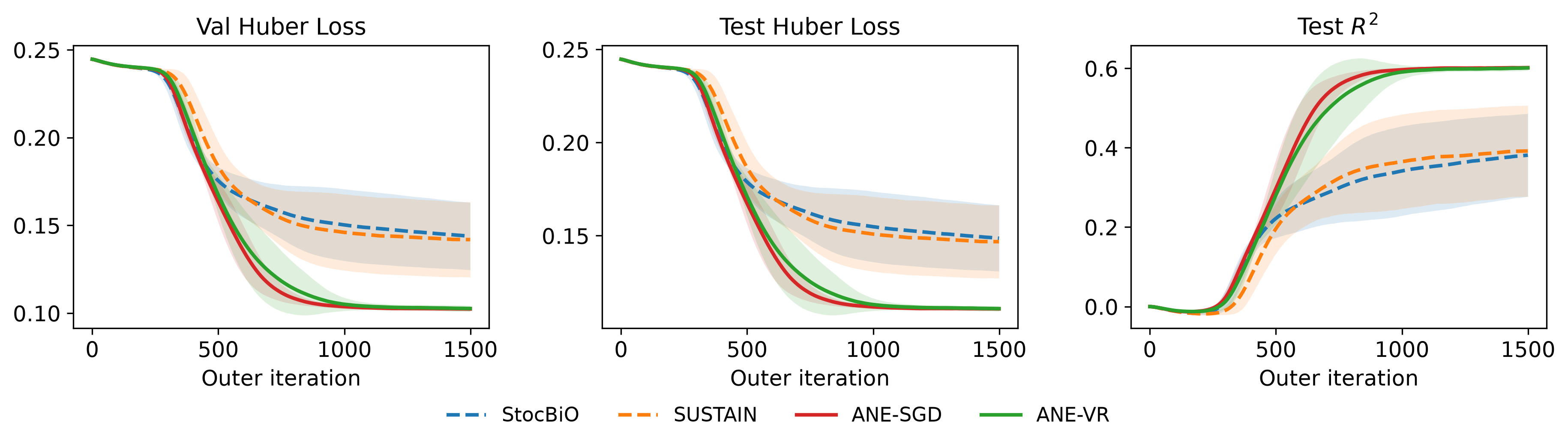}
\caption{{California Housing. $n=20$.}}\label{fig:task1_conv_large}
\end{subfigure}
\par\medskip
\begin{subfigure}{\textwidth}
\includegraphics[width=\textwidth]{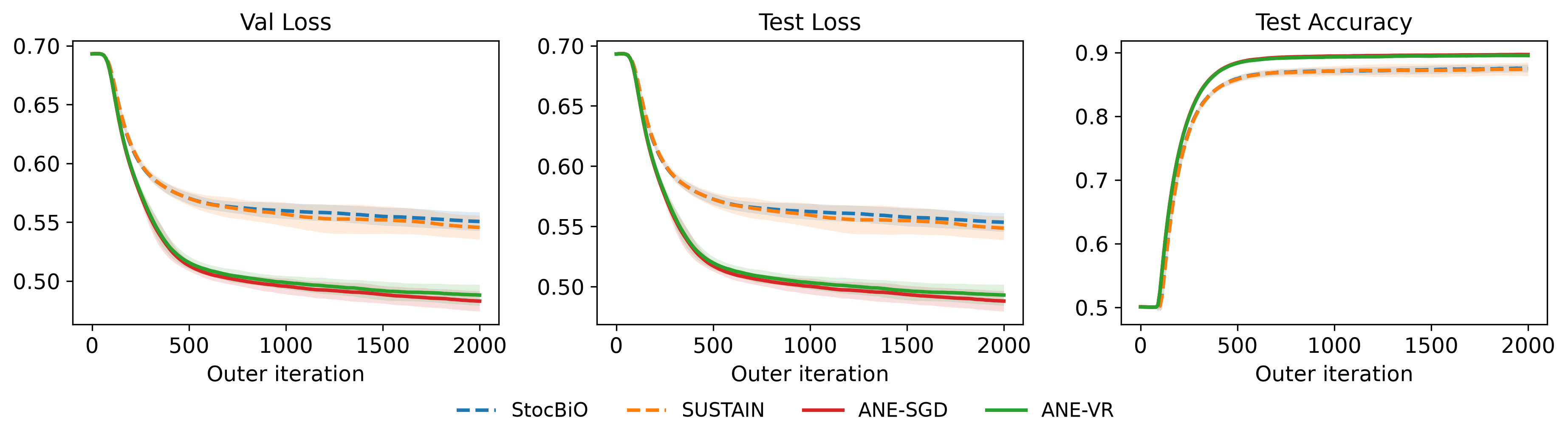}
\caption{{{Fashion-MNIST}. $n=20$.}}\label{fig:task2_conv_large}
\end{subfigure}
\caption{{\textbf{Convergence on bilevel tasks (enlarged).} Each panel: validation (outer) loss, test loss, and the task-specific test metric (mean $\pm$ SD over 20 seeds) vs. outer iteration. Dashed: baselines (StocBiO, SUSTAIN); solid: ANE. {On Housing, Fashion, and Acrobot,} ANE-SGD and ANE-VR separate from baselines early and maintain a consistent gap{; on Fraud, StocBiO has the lowest test loss mid-training and the AUPRC bands overlap}. These comparisons match outer iterations rather than wall-clock cost.}}\label{fig:real_tasks_large}
\end{figure}
\begin{figure}[p]
\ContinuedFloat
\centering
\begin{subfigure}{\textwidth}
\includegraphics[width=\textwidth]{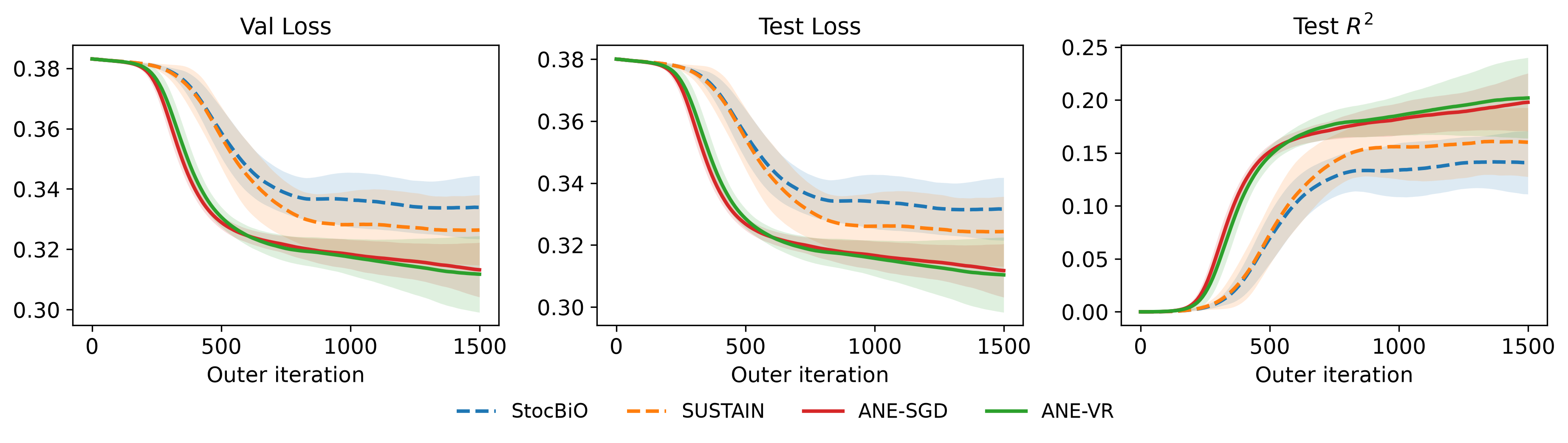}
\caption{{Acrobot. $n=20$.}}\label{fig:task3a_conv_large}
\end{subfigure}
\par\medskip
\begin{subfigure}{\textwidth}
\includegraphics[width=\textwidth]{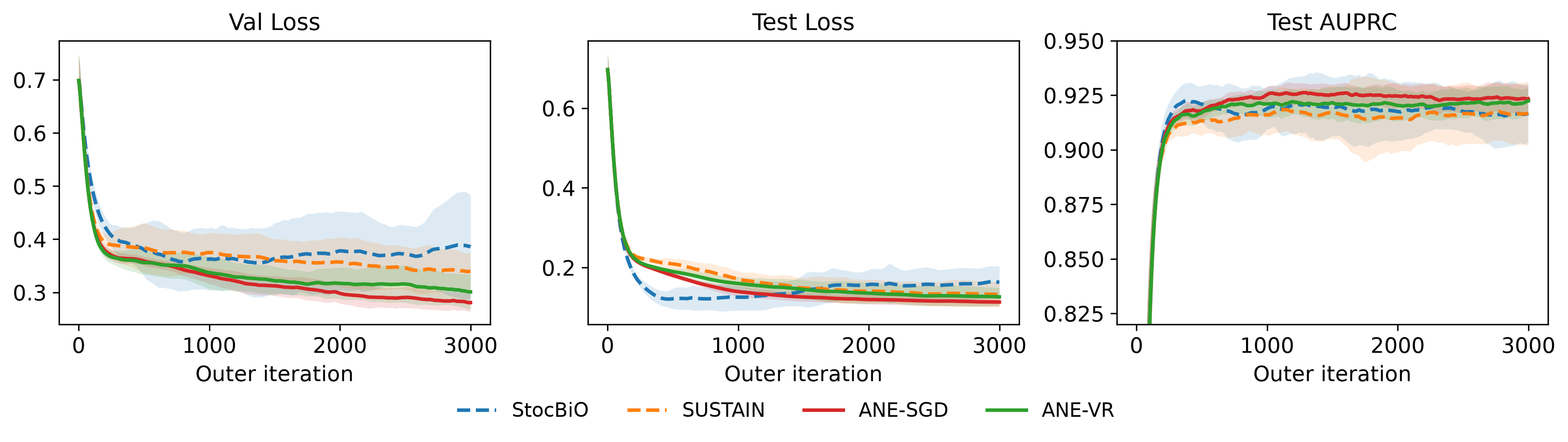}
\caption{{Fraud detection. $n=20$; the validation panel shows the class-balanced outer objective.}}\label{fig:task4_conv_large}
\end{subfigure}
\caption{{\textbf{Convergence on bilevel tasks (continued).} Mean $\pm$ SD over 20 seeds, with the same conventions as the preceding panels.}}
\end{figure}
\clearpage

\end{document}

%% file: tables/tau_ablation.tex
\begin{tabular}{@{}llccccc@{}}
\toprule
Task & Metric & SUSTAIN & $\tau{=}0.5$ & $\tau{=}0.7$ & $\tau{=}0.9$ & $\tau{=}1.0$ \\
\midrule
Housing & Loss $\downarrow$ & $0.147\pm0.020$ & $0.110\pm0.001$ & $0.110\pm0.001$ & $0.110\pm0.001$ & $0.110\pm0.001$ \\
 & $R^2$ $\uparrow$ & $0.393\pm0.116$ & $0.607\pm0.006$ & $0.606\pm0.007$ & $0.605\pm0.007$ & $0.604\pm0.007$ \\
\addlinespace
Fashion & Loss $\downarrow$ & $0.548\pm0.011$ & $0.486\pm0.008$ & $0.487\pm0.009$ & $0.487\pm0.009$ & $0.487\pm0.009$ \\
 & Acc (\%) $\uparrow$ & $87.6\pm1.3$ & $89.7\pm0.3$ & $89.6\pm0.3$ & $89.6\pm0.3$ & $89.6\pm0.3$ \\
\addlinespace
Acrobot & Loss $\downarrow$ & $0.325\pm0.012$ & $0.311\pm0.011$ & $0.310\pm0.011$ & $0.312\pm0.010$ & $0.311\pm0.010$ \\
 & $R^2$ $\uparrow$ & $0.159\pm0.034$ & $0.199\pm0.034$ & $0.202\pm0.034$ & $0.198\pm0.032$ & $0.200\pm0.032$ \\
\addlinespace
Fraud & Loss $\downarrow$ & $0.339\pm0.035$ & $0.284\pm0.018$ & $0.282\pm0.021$ & $0.282\pm0.019$ & $0.281\pm0.017$ \\
 & AUPRC $\uparrow$ & $0.916\pm0.017$ & $0.922\pm0.007$ & $0.923\pm0.007$ & $0.923\pm0.006$ & $0.924\pm0.006$ \\
\bottomrule
\end{tabular}

%% file: tables/controls_matched_legacy.tex
\begin{tabular}{@{}lccc@{}}
\toprule
Estimator & Coord.\ $|\gamma_1|$ & Variance & $\mathcal M_3$ \\
\midrule
Standard & $2.03\pm0.28$ & $7.4\mathrm{e}{-3}$ & $1.6\mathrm{e}{-3}$ \\
Standard (2x) & $1.30\pm0.13$ & $3.5\mathrm{e}{-3}$ & $4.1\mathrm{e}{-4}$ \\
Reference-only ($\tau=0$) & $1.87\pm0.19$ & $1.8\mathrm{e}{-3}$ & $1.8\mathrm{e}{-4}$ \\
ANE ($\tau=1$) & $0.42\pm0.43$ & $3.6\mathrm{e}{-3}$ & $5.2\mathrm{e}{-4}$ \\
\bottomrule
\end{tabular}

%% file: tables/controls_sweep_legacy.tex
\begin{tabular}{@{}lcccc@{}}
\toprule
Multiplier skewness & Standard (2x) $|\gamma_1|$ & ANE $|\gamma_1|$ & Variance, 2x / ANE & $\mathcal M_3$, 2x / ANE \\
\midrule
$0$ (symmetric) & $0.21\pm0.05$ & $0.23\pm0.04$ & $3.6\mathrm{e}{-3}$ / $3.6\mathrm{e}{-3}$ & $3.3\mathrm{e}{-4}$ / $3.3\mathrm{e}{-4}$ \\
$2$ & $0.52\pm0.05$ & $0.18\pm0.09$ & $3.6\mathrm{e}{-3}$ / $3.6\mathrm{e}{-3}$ & $3.3\mathrm{e}{-4}$ / $3.7\mathrm{e}{-4}$ \\
$4$ & $1.30\pm0.13$ & $0.20\pm0.10$ & $3.5\mathrm{e}{-3}$ / $3.6\mathrm{e}{-3}$ & $4.1\mathrm{e}{-4}$ / $4.6\mathrm{e}{-4}$ \\
$6$ & $1.98\pm0.12$ & $0.39\pm0.18$ & $3.7\mathrm{e}{-3}$ / $3.6\mathrm{e}{-3}$ & $5.3\mathrm{e}{-4}$ / $5.6\mathrm{e}{-4}$ \\
\bottomrule
\end{tabular}

%% file: tables/controls_tau_legacy.tex
\begin{tabular}{@{}lcccccc@{}}
\toprule
 & Standard & $\tau=0$ & $\tau=0.25$ & $\tau=0.5$ & $\tau=0.75$ & $\tau=1$ \\
\midrule
Coord.\ $|\gamma_1|$ & $2.03\pm0.28$ & $1.89\pm0.33$ & $1.70\pm0.18$ & $1.29\pm0.47$ & $0.55\pm0.14$ & $0.33\pm0.39$ \\
$\mathcal M_3$ & $1.6\mathrm{e}{-3}$ & $1.9\mathrm{e}{-4}$ & $1.9\mathrm{e}{-4}$ & $2.5\mathrm{e}{-4}$ & $2.9\mathrm{e}{-4}$ & $4.8\mathrm{e}{-4}$ \\
$\mathcal M_3$ below Standard & -- & $88\%$ & $88\%$ & $84\%$ & $81\%$ & $69\%$ \\
Variance & $7.4\mathrm{e}{-3}$ & $1.9\mathrm{e}{-3}$ & $1.9\mathrm{e}{-3}$ & $2.3\mathrm{e}{-3}$ & $2.7\mathrm{e}{-3}$ & $3.6\mathrm{e}{-3}$ \\
\bottomrule
\end{tabular}

%% file: tables/diag_reference_error.tex
\begin{tabular}{@{}lccccc@{}}
\toprule
 & \multicolumn{5}{c}{Checkpoint $t/T$} \\
\cmidrule(l){2-6}
Task & $0$ & $0.25$ & $0.5$ & $0.75$ & $1$ \\
\midrule
Housing & $0.68$ & $0.13$ & $0.47$ & $0.43$ & $0.31$ \\
Fashion & $0.06$ & $1.09$ & $1.99$ & $0.96$ & $1.42$ \\
Acrobot & $0.37$ & $0.08$ & $0.34$ & $0.50$ & $0.45$ \\
Fraud & $0.52$ & $0.96$ & $0.71$ & $0.44$ & $0.74$ \\
\bottomrule
\end{tabular}

%% file: tables/diag_estimators.tex
\begin{tabular}{@{}lcccc@{}}
\toprule
 & \multicolumn{2}{c}{Error moments} & \multicolumn{2}{c}{Skewness} \\
\cmidrule(lr){2-3}\cmidrule(l){4-5}
Estimator & Bias & Variance & $\gamma_1$ along $u$ & Coord.\ $|\gamma_1|$ \\
\midrule
\multicolumn{5}{@{}l}{\textit{Housing} ($m=50$, $t=750$)} \\
\quad Standard & $7.3\mathrm{e}{-3}\pm1\mathrm{e}{-3}$ & $0.014\pm5\mathrm{e}{-3}$ & $+0.18\pm3.78$ & $2.09\pm1.48$ \\
\quad Standard (2x) & $6.6\mathrm{e}{-3}\pm2\mathrm{e}{-3}$ & $9.3\mathrm{e}{-3}\pm6\mathrm{e}{-3}$ & $+0.56\pm2.69$ & $2.62\pm2.14$ \\
\quad ANE & $0.38\pm0.08$ & $17\pm9$ & $+0.91\pm2.18$ & $1.15\pm1.42$ \\
\addlinespace
\multicolumn{5}{@{}l}{\textit{Fashion} ($m=20$, $t=1000$)} \\
\quad Standard & $1.4\pm0.4$ & $70\pm7$ & $+0.06\pm0.22$ & $0.22\pm0.01$ \\
\quad Standard (2x) & $1.5\pm0.1$ & $33\pm2$ & $+0.14\pm0.12$ & $0.17\pm0.02$ \\
\quad ANE & $1.4\pm0.3$ & $203\pm11$ & $-0.13\pm0.07$ & $0.16\pm0.01$ \\
\addlinespace
\multicolumn{5}{@{}l}{\textit{Acrobot} ($m=50$, $t=750$)} \\
\quad Standard & $0.90\pm0.4$ & $46\pm2$ & $-0.05\pm0.05$ & $0.43\pm0.14$ \\
\quad Standard (2x) & $0.61\pm0.2$ & $22\pm2$ & $-0.07\pm0.06$ & $0.43\pm0.10$ \\
\quad ANE & $0.85\pm0.2$ & $61\pm8$ & $-0.06\pm0.24$ & $0.52\pm0.22$ \\
\addlinespace
\multicolumn{5}{@{}l}{\textit{Fraud} ($m=10$, $t=1500$)} \\
\quad Standard & $0.32\pm0.01$ & $1.2\pm0.2$ & $+3.53\pm1.25$ & $5.96\pm0.83$ \\
\quad Standard (2x) & $0.35\pm0.02$ & $1.1\pm0.4$ & $-3.50\pm3.60$ & $3.45\pm0.41$ \\
\quad ANE & $0.56\pm7\mathrm{e}{-3}$ & $5.2\pm1$ & $-2.57\pm1.99$ & $4.64\pm0.81$ \\
\bottomrule
\end{tabular}

%% file: tables/timing.tex
\begin{tabular}{@{}lrcccc@{}}
\toprule
Task & $m$ & StocBiO & SUSTAIN & ANE-SGD & ANE-VR \\
\midrule
Housing & 5 & $1.02$ & $1.69$ ($1.6\times$) & $2.76$ ($2.7\times$) & $3.19$ ($3.1\times$) \\
 & 10 & $1.86$ & $2.96$ ($1.6\times$) & $5.38$ ($2.9\times$) & $7.08$ ($3.8\times$) \\
 & 20 & $3.47$ & $5.20$ ($1.5\times$) & $10.71$ ($3.1\times$) & $13.75$ ($4.0\times$) \\
 & 50 & $8.79$ & $13.23$ ($1.5\times$) & $26.87$ ($3.1\times$) & $33.22$ ($3.8\times$) \\
\addlinespace
Fashion & 5 & $0.93$ & $1.41$ ($1.5\times$) & $2.60$ ($2.8\times$) & $3.35$ ($3.6\times$) \\
 & 10 & $1.69$ & $2.63$ ($1.6\times$) & $5.03$ ($3.0\times$) & $6.15$ ($3.6\times$) \\
 & 20 & $3.23$ & $5.00$ ($1.5\times$) & $10.17$ ($3.1\times$) & $11.29$ ($3.5\times$) \\
 & 50 & $7.79$ & $11.96$ ($1.5\times$) & $25.85$ ($3.3\times$) & $29.49$ ($3.8\times$) \\
\addlinespace
Acrobot & 5 & $1.05$ & $1.73$ ($1.7\times$) & $2.88$ ($2.8\times$) & $3.31$ ($3.2\times$) \\
 & 10 & $1.88$ & $3.01$ ($1.6\times$) & $5.69$ ($3.0\times$) & $7.34$ ($3.9\times$) \\
 & 20 & $3.60$ & $5.37$ ($1.5\times$) & $11.05$ ($3.1\times$) & $13.63$ ($3.8\times$) \\
 & 50 & $8.70$ & $13.18$ ($1.5\times$) & $27.28$ ($3.1\times$) & $33.97$ ($3.9\times$) \\
\addlinespace
Fraud & 5 & $0.92$ & $1.57$ ($1.7\times$) & $2.67$ ($2.9\times$) & $3.01$ ($3.3\times$) \\
 & 10 & $1.72$ & $2.71$ ($1.6\times$) & $5.14$ ($3.0\times$) & $6.82$ ($4.0\times$) \\
 & 20 & $3.33$ & $4.84$ ($1.5\times$) & $10.34$ ($3.1\times$) & $12.74$ ($3.8\times$) \\
 & 50 & $8.03$ & $12.38$ ($1.5\times$) & $25.17$ ($3.1\times$) & $31.98$ ($4.0\times$) \\
\bottomrule
\end{tabular}